\documentclass[peerreview,12pt,comsoc]{IEEEtran}

\usepackage[T1]{fontenc}
\usepackage{amsmath,amsthm,amsfonts,mathtools}
\usepackage[cmintegrals]{newtxmath} 

\IEEEoverridecommandlockouts
\usepackage{cite}
\usepackage{algorithmic}
\usepackage{graphicx}
\usepackage{pspicture}
\usepackage{textcomp}
\def\BibTeX{{\rm B\kern-.05em{\sc i\kern-.025em b}\kern-.08em
    T\kern-.1667em\lower.7ex\hbox{E}\kern-.125emX}}

\mathtoolsset{showonlyrefs=true} 
\usepackage{bbm}

\mathchardef\Re="023C
\mathchardef\Im="023D
\usepackage{breqn}
\usepackage{float}
\usepackage{caption}
\usepackage{subcaption}
\usepackage{gensymb}
\usepackage{verbatim}
\usepackage{euscript}

\usepackage{standalone}

\newcommand*\conj[1]{\overline{#1}}
\DeclareMathOperator{\E}{\mathbb{E}}

\newcommand{\rem}[1]{}

\newcommand{\bre}{\begin{equation}}
\newcommand{\ere}{\end{equation}}

\newcommand{\ee}\]
\newcommand{\bfg}{\begin{figure}[hbtp]}
\newcommand{\efg}{\end{figure}}
\newcommand{\bver}{\begin{verbatim}}
\newcommand{\ever}{\end{verbatim}}
\newcommand{\bit}{\begin{itemize}}
\newcommand{\eit}{\end{itemize}}
\newcommand{\ben}{\begin{enumerate}}
\newcommand{\een}{\end{enumerate}}

\newcommand{\bpsi}{\boldsymbol\psi}

\newcommand{\btheta}{\boldsymbol\theta}

\newcommand{\given}{\: | \:}

\newcommand{\bphi}{{\mathbf \Phi}}

\newcommand{\bg}{{\bf{g}}}
\newcommand{\bG}{{\bf{G}}}

\newcommand{\bc}{{\bf c}}

\newcommand{\bm}{{\bf m}}

\newcommand{\baa}{\begin{eqnarray*}}
\newcommand{\eaa}{\end{eqnarray*}}

\newcommand{\bh}{{\bf h}}

\newcommand{\bu}{{\bf u}}

\newcommand{\bw}{{\bf w}}

\newcommand{\bx}{{\bf x}}
\newcommand{\by}{{\bf y}}

\newcommand{\cL}{{\cal L}}

\newcommand{\cH}{{\cal H}}

\newcommand{\cJ}{{\cal J}}

\newcommand{\cC}{{\mathcal{C}}}
\newcommand{\cI}{{\mathcal{I}}}

\newcommand{\cN}{{\mathcal{N}}}
\newcommand{\cG}{\mathcal{G}}

\newcommand{\bhh}{\hat{\bf h}}

\newcommand{\defined}{\triangleq}

\def\defined{\: {\stackrel{\scriptscriptstyle \Delta}{=}} \: }

\newfont{\boldlarge}{msbm10 scaled 1100}

\newlength{\tmpbigbar}

\usepackage[utf8]{inputenc}
\usepackage[ruled,longend]{algorithm2e}
\usepackage{textgreek}

\usepackage{varwidth}
\usepackage{tikz}
\usepackage{textcomp}
\usetikzlibrary{shapes,arrows}
\usetikzlibrary{positioning}

\usetikzlibrary{arrows,
                chains,
                decorations.markings,
                shadows, shapes.arrows,shapes}

\usetikzlibrary{arrows.meta,
	chains,
	decorations.pathreplacing,
	calligraphy,
	arrows,
	decorations.markings,
	shadows, shapes.arrows,shapes, positioning, fit
}
\usepackage{array}


\begin{document}

\title{Unsupervised Linear and Nonlinear Channel Equalization and Decoding using Variational Autoencoders}

\author{Avi Caciularu and David~Burshtein,~\IEEEmembership{Senior Member,~IEEE}
	\thanks{This research was supported by the Israel Science Foundation (grant no. 1868/18), and by the Yitzhak and Chaya Weinstein Research Institute for Signal Processing.}
	\thanks{A.\ Caciularu is with the school of Electrical Engineering, Tel-Aviv University, Tel-Aviv 6997801, Israel (email: avi.c33@gmail.com).}
	\thanks{D.\ Burshtein is with the school of Electrical Engineering, Tel-Aviv University, Tel-Aviv 6997801, Israel (email: burstyn@eng.tau.ac.il).}
}


\maketitle \setcounter{page}{1}

\begin{abstract}
A new approach for blind channel equalization and decoding, variational inference, and variational autoencoders (VAEs) in particular, is introduced. We first consider the reconstruction of uncoded data symbols transmitted over a noisy linear intersymbol interference (ISI) channel, with an unknown impulse response, without using pilot symbols. We derive an approximate maximum likelihood estimate to the channel parameters and reconstruct the transmitted data. We demonstrate significant and consistent improvements in the error rate of the reconstructed symbols, compared to existing blind equalization methods such as constant modulus, thus enabling faster channel acquisition. The VAE equalizer uses a convolutional neural network with a small number of free parameters. These results are extended to blind equalization over a noisy nonlinear ISI channel with unknown parameters. We then consider coded communication using low-density parity-check (LDPC) codes transmitted over a noisy linear or nonlinear ISI channel. The goal is to reconstruct the transmitted message from the channel observations corresponding to a transmitted codeword, without using pilot symbols. We demonstrate improvements compared to the expectation maximization (EM) algorithm using turbo equalization. Furthermore, unlike EM, the computational complexity of our method does not have exponential dependence on the size of the channel impulse response.  
\end{abstract}


\section{Introduction}
Deep learning methods have recently been considered in various communication problems.
For example, in \cite{nachmani2016learning,nachmani2018deep,tenbrink,cammerer2017scaling, raviv2020perm2vec} deep learning methods were considered to solve the problem of channel decoding,
in \cite{AutoencoderComm} an autoencoder for short blocklength end-to-end communications was proposed,
and in \cite{samuel2017deep} deep learning was used for MIMO detection.
Various authors have considered deep learning-based channel equalization, separately or jointly with the decoding task.
In \cite{farsad2018neural} deep learning-based detection algorithms were used when the channel model is unknown.
In \cite{liang2018iterative} the authors show improved decoding of low-density parity-check (LDPC) codes, by augmenting the belief propagation (BP) algorithm with a convolutional neural network. In \cite{ye2018channel,o2019approximating}, generative adversarial networks (GANs) were proposed to model channel effects in end-to-end communication systems.

In this work, we consider transmission over a noisy intersymbol interference (ISI) channel with an unknown impulse response. The ISI channel is useful to model various communication scenarios, such as multipath in wireless channels \cite{proakis2001digital}. Unlike the works that were mentioned above, we do not assume the availability of a pilot signal to learn the unknown channel. The motivation is that pilot symbols reduce the communication throughput, especially when the communication environment is changing rapidly. Nor do we assume any prior knowledge on the ISI channel. Hence, decision directed equalization cannot be used. Instead, unsupervised, blind channel equalization is proposed. The method is unsupervised in the sense that we cannot assume the availability of channel output samples corresponding to known transmitted symbols.
For uncoded transmitted data, the standard approach for blind channel equalization is the constant modulus algorithm (CMA) \cite{godard1980self,treichler1983new,johnson1998blind}. Blind neural network-based algorithms using the constant modulus (CM) criterion were proposed in \cite{you1998nonlinear}.
The maximum likelihood (ML) criterion has also been considered for blind channel equalization \cite{ghosh1992maximum,tong1998multichannel,cirpan1999maximum,kaleh1994joint,gunther2007generalized} (and references therein).
Combined blind equalization and decoding using expectation maximization (EM) for the case where the transmitted data is coded was considered in \cite{wang2001blind,1510991}.
The proposed solutions use the EM algorithm \cite{dempster1977maximum} or an approximate EM, which requires an iterative application of the Bahl, Cocke, Jelinek, and Raviv (BCJR) algorithm \cite{bahl1974optimal} or the Viterbi algorithm. Hence, the complexities of these algorithms (both execution time and memory consumption) are exponential in the channel memory size, which may be prohibitive. In fact, for a channel impulse response of size $M$, and signal constellation of size $q$ (e.g., for BPSK $q=2$ and for QPSK $q=4$), the number of states in the BCJR trellis is $q^{M-1}$.

In our work we also consider nonlinear channels, where the nonlinear distortion may be due to the presence of amplifiers, converters and mixers in transmitters and receivers.
Channel equalization is applied to overcome these effects and reconstruct the signal before using the channel decoder.
Some authors \cite{mitchinson2002digital,patra2009nonlinear,olmos2010joint,ye2017initial,8491056} have considered the nonlinear channel equalization problem under a supervised learning setup, which requires pilot signals for training the equalizer.

In the first part of this work, which was initially presented in \cite{8403666}, we present a new approach for unsupervised blind channel equalization of uncoded data, transmitted over a noisy ISI channel with an unknown impulse response, without the availability of pilot symbols.
The method uses variational inference, e.g. \cite{bishop2006pattern,blei2017variational}, and variational autoencoders (VAEs) in particular \cite{kingma2013auto,rezende2014stochastic} as a means to obtain an approximate maximum likelihood estimate to the channel parameters.
VAEs are widely used in the literature of deep learning for unsupervised and semi-supervised learning, and as a generative model.
We demonstrate significant and consistent improvements in the error rate of the reconstructed symbols, compared to existing blind equalization methods such as constant modulus, thus enabling faster channel acquisition.
In fact, for the channels that were examined, the performance of the new blind VAE equalizer (VAEE) was close to the performance of a non-blind adaptive linear minimum mean square error (MMSE) equalizer \cite{gong2010adaptive}.
Furthermore, unlike the ML-based blind equalization methods in the literature \cite{ghosh1992maximum,tong1998multichannel,cirpan1999maximum,kaleh1994joint,gunther2007generalized}, the computational complexity of our approach is not exponential in the channel memory size.
In fact, the VAEE uses a convolutional neural network with a very small number of free parameters.
We report on the number of iterations that the VAEE requires later on in the paper, and we also comment on the computational complexity of each iteration.
These results are extended to unsupervised blind equalization over a noisy nonlinear ISI channel with unknown parameters. We then consider coded communication using LDPC codes transmitted over a noisy linear or nonlinear ISI channel. Here the goal is to decode a transmitted codeword when the only data available for decoding are the channel observations corresponding to this unknown transmitted codeword.
Our method is shown to be superior to other methods for blind channel equalization, e.g., EM using Turbo equalization \cite{douillard1995iterative,koetter2004turbo}, where only a genie-aided version worked well for some channels in the coded data setup. In particular, for nonlinear channels, the EM algorithm must ignore the unknown nonlinearity while our method can model it using a VAE-decoder neural network.
Furthermore, unlike EM, the computational complexity of our method does not have exponential dependence on the size of the channel impulse response.

Our main contribution in this paper can be summarized as follows:
\begin{itemize}
\item We present a new method for blind channel equalization using variational inference, and VAEs in particular.
\item We extend the method to the case of coded communications.
\item Both for linear and for nonlinear channels we demonstrate improvements compared to the baseline methods for blind channel equalization.
\end{itemize}

The rest of this paper is organized as follows.
In Section \ref{sec:problem_setup} we present the problem setup considered in this paper. In Section \ref{sec:lin_chan} we present our proposed solution for uncoded data transmitted over a noisy linear ISI channel. In Section \ref{sec:nonlin_chan} we consider the same setup for noisy nonlinear channels. In Section \ref{coded_data} we consider LDPC coded data transmitted over noisy (linear or nonlinear) ISI channels, and present joint blind equalization and decoding using our method. In Section \ref{sec:sim} we present simulation results. Finally, Section \ref{sec:conclusion} concludes this paper.

\section{Problem setup} \label{sec:problem_setup}
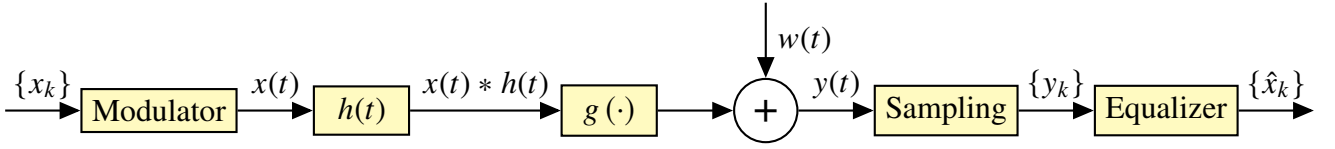
\begin{figure}[]
\centering
\scalebox{1}{\tikzset{%
	block/.style    = {draw, thick, rectangle, minimum height = 1em,
		minimum width = 3em, fill=yellow!30!white,},
	sum/.style      = {draw, circle, node distance = 2cm}, 
	input/.style    = {coordinate}, 
	output/.style   = {coordinate} 
}
\newcommand{\suma}{\Large$+$}

\begin{tikzpicture}[auto, thick, node distance=2cm, >=triangle 45]
	\draw
	node at (0,0){}
	node [input, name=input1] {} 
	node [block, right = 1cm of input1] (bla) {Modulator}
	node [block, right = 1cm of bla] (inte2) {$h(t)$}
	node [block, right = 2cm of inte2] (inte3) {$g\left(\cdot\right)$}
	node [sum, right = 1.0cm of inte3] (suma1) {\suma}
	node [input, name=input2, above = 1cm of suma1] {} 
	node [block, right = 1cm of suma1] (inte1) {Sampling}
	node [block, right = 1cm of inte1] (Q1) {Equalizer}
	node [output, name=output1, right = 1cm of Q1] {};
	\draw[->](input1) -- node {$\{x_k\}$}(bla);
	\draw[->](bla) -- node {$x(t)$}(inte2);
	\draw[->](inte2) -- node {$x(t)*h(t)$} ++(2.65cm,0) (inte3);
	\draw[->](inte3) -- node {} (suma1);
	\draw[->](suma1) -- node {$y(t)$} (inte1);
	\draw[->](inte1) -- node {$\{y_k\}$} (Q1);
	\draw[->](Q1) -- node {$\{\hat x_k\}$} (output1);
	\draw[->](input2) -- node {$w(t)$} (suma1);
\end{tikzpicture}}
\caption{End-to-end communication system model.}
\label{fig:sys}
\end{figure}

The communication channel is modeled as a convolution of the input, $\{ x_k \}$, with some causal, finite impulse response (FIR), time invariant filter, $\bh=(h_0,h_1,\ldots,h_{M-1})$, of size $M$, followed by the possibly nonlinear mapping $g(\cdot)$ and the addition of white Gaussian noise
\begin{equation}
y_n = g \left( \sum_{k} x_k h_{n-k} \right) + w_n
\label{eq:sys}
\end{equation}
This is the equivalent model of $\{ y_k \}$ in the end-to-end communication system shown in Fig. \ref{fig:sys}. In this figure, $x(t)$ is the modulated analog signal, $h(t)$ is the analog channel impulse response, and $y(t)=g(x(t)*h(t))+w(t)$ is the received noisy channel output signal. The samples $\{ y_k \}$ are obtained by sampling $y(t)$ at the symbol rate.

The equalizer in Fig. \ref{fig:sys} reconstructs an estimate of the transmitted symbol sequence, $\{\hat{x}_k\}$. Now, suppose that we observe a finite window of measurements $\by\defined (y_0,y_1,\ldots,y_{N-1})$.
For clarity of presentation, we assume that the input signal is causal ($x_k=0$ for $k<0$). We refer to this assumption later.
Equation \eqref{eq:sys} can be written compactly for the measurements collected in $\by$ as
\begin{equation}
\by = \bg(\bx * \bh) + \bw
\label{eq:conv}    
\end{equation}
where $\bx=(x_0,x_1,\ldots,x_{N-1})$ is the transmitted message, and $\bw=(w_{0},w_1,\ldots,w_{N-1})$ is an i.i.d. sequence of additive white Gaussian noise. The function $\bg()$ in \eqref{eq:conv} is defined by $\bg(\bu)\defined (g(u_0),\ldots,g(u_{N-1}))$ for the input vector $\bu=(u_0,\ldots,u_{N-1})$ (i.e., the scalar function $g(\cdot)$ is applied component-wise on $\bx * \bh$).
Note that by setting $g(x)=x$ the model degenerates to a noisy linear ISI channel.
In this paper we consider both BPSK and QPSK modulation, although the derivation can be extended to other constellations.

Our goal in this paper is to design the equalizer that reconstructs the transmitted sequence $\bx$. The design utilizes only the channel observations $\by$, i.e., without knowing the channel parameters, including the impulse response $\bh$, the nonlinear function $g(\cdot)$ and the noise variance. This is an unsupervised blind channel equalization problem where pilot signals are not available.

We will discuss both the case of uncoded data, $\{x_k\}$ (or when coding information is not used), and the case of coded data.

\subsection{BPSK modulation} 
For BPSK modulation, $x_k = \pm 1$.
We assume a uniformly distributed transmitted sequence, so that for all $\bx\in\{-1,1\}^N$ the probability of $\bx$ is given by
\begin{equation}
p(\bx)= 2^{-N}
\label{eq:px_BPSK}
\end{equation}
The noise, $\bw$, is a sequence of independent identically distributed (i.i.d) Gaussian random variables with variance $\sigma_w^2$. 
Given $\bx$, $\by$ is a vector of statistically independent, normally distributed components.
The conditional density function of $\by$ is $\cN(\bg(\bx*\bh),\sigma^{2}_{w} I_N)$.
Thus, for $\btheta \defined \left\{ \bh, g(\cdot), \sigma_w^2 \right\}$, the conditional density of $\by$ given $\bx$ can be expressed as 
\bre
p_{\btheta} (\by \given \bx) 
=
\frac{1}{\left(2\pi \sigma^{2}_{w} \right)^{N/2}} \cdot
e^{-\left\lVert \by-\bg(\bx*\bh) \right \rVert^{2} / [2\sigma^{2}_{w}]} 
\label{eq:pyx_BPSK}
\ere

\subsection{QPSK modulation} 
For QPSK modulation, $x_k = \pm 1 \pm j$, and the above vectors can be written as combinations of real ($I$) and imaginary ($Q$) components, so that, $\bx=\bx^{I}+j\cdot \bx^{Q}$, $\bh=\bh^{I}+j\cdot \bh^{Q}$ and $\by=\by^{I}+j\cdot \by^{Q}$.
We assume a uniformly distributed transmitted sequence, so that for all valid $\bx$ the probability of $\bx$ is given by
\begin{equation}
p(\bx)= p(\bx^I) p(\bx^Q) = 2^{-2N}
\label{eq:px_QPSK}
\end{equation}
Each element of the i.i.d noise sequence, $\bw$, is complex Gaussian with statistically independent real and imaginary components, each with variance $\sigma_w^2/2$. 
Given $\bx$, $\by^I$ and $\by^Q$ are statistically independent, normally distributed.
The conditional density function of $\by^I$ is $\cN(\Re \left(\bg(\bx*\bh)\right),(\sigma^{2}_{w}/2) I_N)$. The conditional density function of $\by^Q$ is $\cN(\Im \left(\bg(\bx*\bh)\right),(\sigma^{2}_{w}/2) I_N)$.
Thus, for $\btheta \defined \left\{ \bh, g(\cdot), \sigma_w^2 \right\}$, the conditional density of $\by$ given $\bx$ can be expressed as
\begin{align}
p_{\btheta} (\by \given \bx) 
&= p_{\btheta} \left( \by^I \given \bx \right)
p_{\btheta} \left( \by^Q \given \bx \right) \nonumber\\
&=
\frac{1}{\left(\pi \sigma^{2}_{w} \right)^{N}} \cdot
e^{-\left\lVert \by-\bg(\bx*\bh) \right \rVert^{2} / \sigma^{2}_{w}} 
\label{eq:pyx_QPSK}
\end{align}

\section{Proposed method for a noisy linear ISI channel} \label{sec:lin_chan}
In this section we consider the uncoded case, where coding information in $\{x_k\}$ is not used. We also assume that $g(x) = x$ (see Fig. \ref{fig:sys}) so that the channel is linear. In the next section we consider the case of a nonlinear channel. We start with the case of QPSK modulation, and then note how the results simplify for BPSK modulation.

\subsection{QPSK modulation} \label{sec:lin_chan_QPSK}
We propose using ML estimation of the channel impulse response, $\bh$, and noise variance, $\sigma_w^2$. That is, we search for the vector $\btheta=(\bh,\sigma_w^2)$ that maximizes\footnote{The default base of the logarithms in this paper is $e$.} $\log p_\btheta(\by)$. The ML estimate has strong asymptotic optimality properties, and in particular asymptotic efficiency \cite{tong1998multichannel}. For the CMA criterion, on the other hand, one can only claim asymptotic consistency \cite{shalvi1990new}.
However, applying the exact ML criterion to our problem is very difficult since $p_\btheta(\by)$ should first be expressed as
\begin{equation*}
p_\btheta(\by) = \sum_{\bx} p(\bx) p_\btheta(\by \given \bx)
\end{equation*}
where we sum over all $2^{2N}$ possible input sequences $\bx$ and where $p(\bx)$ is given by \eqref{eq:px_QPSK}.
Nevertheless, for this kind of problems, it has been shown in various applications that it is possible to simplify the estimation problem by using the variational inference approach for ML estimation, e.g. \cite{bishop2006pattern,blei2017variational}.
By the variational inference approach, instead of directly maximizing $p_\btheta(\by)$ over $\btheta$, one maximizes iteratively a variational lower bound, also called evidence lower bound (ELBO), as follows. It can be shown, e.g., \cite{bishop2006pattern,blei2017variational,kingma2013auto} that
\begin{align}
\log p_{\btheta} (\by) &\geq \E_{q_{\bphi}(\bx\given\by)}
\left[-\log q_{\bphi}(\bx\given\by)+\log p_{\btheta} (\bx,\by)\right]
\\
&=
\underbrace{-D_{KL} \left[ q_{\bphi}(\bx\given\by) || p(\bx)\right]}_{A}
+
\underbrace{\E_{q_{\bphi}(\bx\given\by)}
\left[\log p_{\btheta}(\by\given\bx)\right]}_{B} \defined -\cL\left(\btheta,\bphi, \by \right)
\label{eq:elbo}
\end{align}
where $D_{KL}[\cdot || \cdot]$ denotes the Kullback Leibler distance between two density functions, and $q_\bphi(\bx\given\by)$ is an arbitrarily parametrized (by $\bphi$) conditional density function.
Now, instead of directly maximizing $p_\btheta(\by)$, one maximizes the lower bound $-\cL \left(\btheta,\bphi,\by \right)$ over $\btheta$ and $\bphi$ jointly.
Following \cite[Fig. 1]{kingma2013auto} , Fig. \ref{fig:VAE_graph} shows a directed graphical model that describes the communication channel and the network structure used for inference.
\usetikzlibrary{bayesnet}
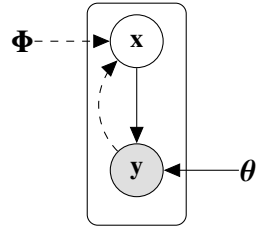
\begin{figure}[]
\begin{center}
\begin{tikzpicture}[scale=1, transform shape]
\node[obs] (y1) {$\mathbf{y}$};
\node[latent, above=of y1] (x1) {$\mathbf{x}$};
\node[const, left=of x1] (phi1) {$\mathbf{\bphi}$};
\node[const, right=of y1] (theta1) {$\mathbf{\btheta}$};
\edge [dashed] {phi1} {x1};
\edge {theta1} {y1};
\draw (y1) edge[out=135,in=225,->,dashed] (x1);
\edge {x1} {y1};
\plate [xscale=1.3] {} {(y1)(x1)} {} ;
\end{tikzpicture}
\end{center}
\caption{A directed graphical model describing the communication channel and the network structure used for inference (following \cite[Fig. 1]{kingma2013auto}). The generative model $p(\bx)p_\btheta(\by|\bx)$ is denoted by solid lines. The variational approximation $q_\bphi(\bx|\by)$ to the posterior $p_\btheta(\bx|\by)$ is denoted by dashed lines.}
\label{fig:VAE_graph}
\end{figure}

In fact, it can be shown \cite{bishop2006pattern,blei2017variational,kingma2013auto} that by searching over $\btheta$ and \emph{all} possible conditional densities $q(\bx\given\by)$, one obtains the ML estimate of $\btheta$.
Furthermore, if $p_\btheta(\bx\given\by)$ can be computed analytically, then, by setting $q_\bphi(\bx|\by)=p_{\btheta'}(\bx\given\by)$, where $\btheta'$ is the value of $\btheta$ at the beginning of the iteration (there is no unknown $\bphi$ in this case) yields the EM algorithm \cite{bishop2006pattern,blei2017variational}.

Typically, when using the VAE approach, both $p_\btheta(\by\given\bx)$ and $q_\bphi(\bx\given\by)$ are implemented using neural networks.
A probabilistic \emph{VAE-encoder} network implements $q_\bphi(\bx\given\by)$, and a probabilistic \emph{VAE-decoder} network implements $p_\btheta(\by\given\bx)$. The unobserved variables $\bx$ provide a latent representation of the data $\by$ \cite{kingma2013auto}.
In our problem, for a noisy linear ISI channel, $p(\bx)$ is given in \eqref{eq:px_QPSK}, and the VAE-decoder, $p_\btheta(\by\given\bx)$, is given in \eqref{eq:pyx_QPSK} (with $g(x)=x$).
Hence, for a noisy linear ISI channel, unlike a standard VAE, we do not need a neural network to implement the VAE-decoder. However, we are using a neural network to implement the VAE-encoder, $q_{\bphi}(\bx\given\by)$, which acts as an equalizer in our problem. Given the channel output sequence, $\by$, we first use it to estimate the VAE parameters $\btheta$ and $\bphi$. Then we obtain an estimated bit sequence, $\bx$, using the hard decoded output of the VAE-encoder (equalizer), $q_\bphi(\bx\given\by)$.
We use the following model,
\begin{equation*}
q_{\bphi}(\bx\given\by)
=
\prod_{j=0}^{N-1} q_{\bphi,j}(x_{j}|\by)
=
\prod_{j=0}^{N-1} q_{\bphi,j}^{I}(x_{j}^{I}\given\by)q_{\bphi,j}^{Q}(x_{j}^{Q}\given\by) 
\end{equation*}
Recalling that $x_j^I \in \{-1,1\}$ and $x_j^Q \in \{-1,1\}$, this is a multivariate Bernoulli distribution with statistical independence between components. 
Denoting by
\begin{align*}
	q_{\bphi,j}^{I}(\by) &\defined q_{\bphi,j}^I(X_{j}^{I}=1|\by)\\
	q_{\bphi,j}^{Q}(\by) &\defined q_{\bphi,j}^Q(X_{j}^{Q}=1|\by)
\end{align*}
the conditional probabilities (under the variational model) of the events $X_{j}^{I}=1$ and $X_{j}^{Q}=1$ given $\by$, we have
\begin{equation*}
q_{\bphi}(\bx\given\by)
=
\prod_{j=0}^{N-1} 
\left( q_{\bphi,j}^{I}(\by) \right)^{(1+x_j^I)/2} \left( 1-q_{\bphi,j}^{I}(\by) \right)^{(1-x_j^I)/2}
\left( q_{\bphi,j}^{Q}(\by) \right)^{(1+x_j^Q)/2} \left( 1-q_{\bphi,j}^{Q}(\by) \right)^{(1-x_j^Q)/2}
\end{equation*}

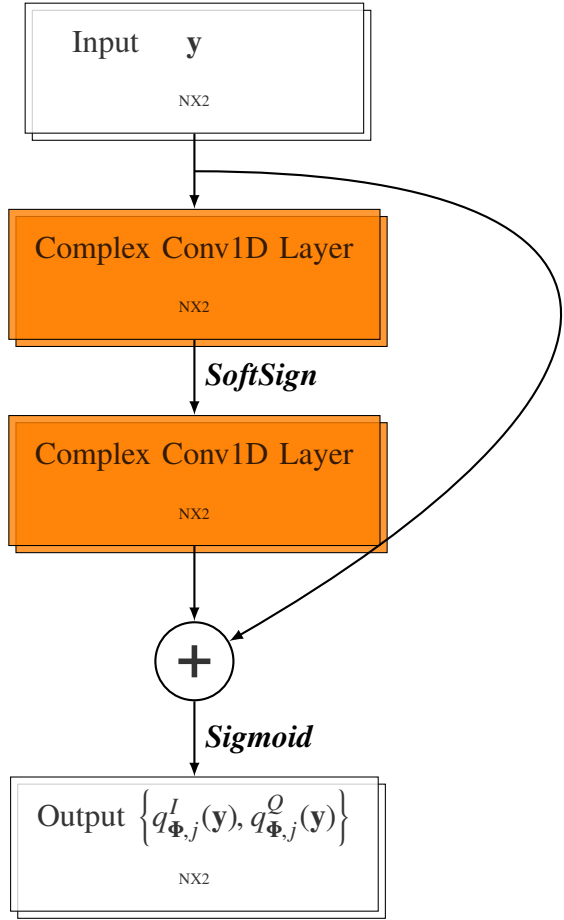
\begin{figure}[]
\centering
   \scalebox{1}{\begin{tikzpicture}
\begin{scope}[start chain = going below,
every node/.append style={on chain,fill opacity=0.8,draw,
	join},every join/.style={thick,-latex},
cs/.style={minimum width=4.5cm,copy shadow={shadow scale=1, shadow xshift=0.5ex, shadow yshift=-0.5ex}}
]
\node[fill=white,cs] (N1) {\begin{tabular}{p{1cm}cp{1cm}}
	Input & $\by$ &\\
	& \tiny NX2 & 
	\end{tabular}};
\node[fill=orange,cs] (N2) {\begin{tabular}{c}
	Complex Conv1D Layer\\
	\tiny NX2 
	\end{tabular}};
\node[fill=orange,cs] (N3) {\begin{tabular}{c}
	Complex Conv1D Layer\\
	\tiny NX2 
	\end{tabular}};
\node[circle,minimum size=8mm,thick,font=\Huge\bfseries] (C) {+};
\node[fill=white,cs] (N4) {\begin{tabular}{c}
	Output $ \left\{ q_{\bphi,j}^{I}(\by), q_{\bphi,j}^{Q}(\by) \right\}$\\
	\tiny NX2 
	\end{tabular}};
\end{scope}  
\path (N1) -- (N2) coordinate[pos=0.5] (aux);
\draw[thick,-latex] (aux) to[out=0,in=30,looseness=2.7] (C);
\path (N2) -- (N3) node[pos=0.5,right,font=\bfseries\itshape]{SoftSign};
\path (C) -- (N4) node[pos=0.5,right,font=\bfseries\itshape]{Sigmoid};
\end{tikzpicture}}
   \caption{Our equalizer's architecture (VAE-encoder network) using complex convolutional residual network. Each convolution output is listed as $\text{Width}\times \#\text{Channels}$.}
\label{fig:architecture}
\end{figure}

In our implementation of the VAE-encoder, we used a convolutional neural network to implement $q_{\bphi,j}^{I}(\by)$ and $q_{\bphi,j}^{Q}(\by)$. The network has complex convolutional layers, each with two output channels, corresponding to the real and imaginary parts of the convolution as in \cite{trabelsi2018deep,o2016radio}. The input and output layers are also separated to two channels corresponding to the real and imaginary components of the input, $\by$, and the output probabilities, $q$. The convolutional layers are both one dimensional (1D) as in \cite{o2016radio}, and with a residual connection as in \cite{resnet}.
The nonlinear activation function of the first layer is a SoftSign function defined by 
$
f\left(x\right)=\frac{x}{|x|+1}
$,
which, in our experiments, proved to converge faster than other functions such as LeakyReLU and Tanh. The nonlinear activation function of the second layer is a sigmoid function, that ensures that the outputs are in $[0,1]$, and so they represent valid probability values. 
Note that each convolutional layer uses only one filter. Using more than one filter did not improve results.
Our VAE-encoder neural network, implementing $\{ q_{\bphi,j}^{I}(\by), q_{\bphi,j}^{Q}(\by) \}$, is depicted in Fig. \ref{fig:architecture}.

We now derive an explicit expression for the loss $\cL\left(\btheta,\bphi, \by \right)=-A-B$ (see Eq. \eqref{eq:elbo}) that needs to be minimized with respect to both $\btheta$ and $\bphi$ (alternatively, $-\cL\left(\btheta,\bphi, \by \right)$ needs to be maximized).

For the term $A$ we have
\begin{align}
A &= \sum_{\bx} q_{\bphi}(\bx|\by) \cdot \left(  \log p(\bx) - \log q_{\bphi}(\bx|\by) \right)\\
  &= \sum_{\bx} q_{\bphi}(\bx|\by) \cdot \left(  -2N\log 2 - \log q_{\bphi}(\bx|\by) \right)\\
  &= -2N\log 2 + \cH \left[ q_{\bphi}(\bx|\by) \right]
  \label{eq:A}
\end{align}
where $\cH \left[ q_{\bphi}(\bx|\by) \right]$, the entropy of $q_{\bphi}(\bx|\by)$, is given by
\begin{align}
\MoveEqLeft[4]
\mathcal{H}\left[ q_{\bphi}(\bx|\by) \right]
=
\mathcal{H}\left[ \prod_{j=0}^{N-1}q_{\bphi,j}(x_{j}|\by) \right]
=\sum_{j=0}^{N-1}  \cH \left[ q_{\bphi,j}(x_{j}|\by) \right]
\\
=-\sum_{j=0}^{N-1} &\left[
q_{\bphi,j}^{I}(\by) \log q_{\bphi,j}^{I}(\by) + \left(1 - q_{\bphi,j}^{I}(\by) \right) \log \left(1 - q_{\bphi,j}^{I}(\by) \right)
\right. \\
&\left. + q_{\bphi,j}^{Q}(\by) \log q_{\bphi,j}^{Q}(\by) + \left(1 - q_{\bphi,j}^{Q}(\by) \right) \log \left(1 - q_{\bphi,j}^{Q}(\by) \right) \right]
\label{eq:Hq}
\end{align}
For the term $B$ we have
\begin{align}
B&=\E_{q_{\bphi}(\bx|\by)}\left[-N\log \left(\pi \sigma_{w}^{2}\right) - \frac{1}{\sigma_{w}^{2}}\left\lVert \by-\bx*\bh \right \rVert^{2} \right]\\
&=-N\log \left( \pi \right) -N\log \left(\sigma_{w}^{2}\right)\\
&\quad - \frac{1}{\sigma_{w}^{2}}\cdot \underbrace{\E_{q_{\bphi}(\bx|\by)}\left[\left\lVert \by-\bx*\bh \right \rVert^{2} \right]}_{C}
\label{eq:B}
\end{align}
We now compute the term $C$ (and hence $B$) analytically. This is possible due to the special structure of the problem, where $p_{\btheta} (\by \given \bx)$ is given by the closed form expression \eqref{eq:pyx_QPSK}, with $g(x)=x$.
First, by the definition of $C$ we have,
\begin{align}
C
&=
\sum_{n=0}^{N-1}
\left[\left|y_{n}\right|^{2}-2\Re \left(y_{n}\sum_{k=0}^{N-1}\E_{q_{\bphi}(\bx|\by)}\left\{\conj{x_{k}}\right\}\conj{h_{n-k}}\right) \right.\\
&\qquad\quad
\left.
+\sum_{k,l=0}^{N-1}\E_{q_{\bphi}(\bx|\by)}\left\{x_{k}\conj{x_{l}}\right\}h_{n-k}\conj{h_{n-l}} 
\right]
\label{eq:C}
\end{align}
where $\overline{(\cdot)}$ denotes the complex conjugate. Now, 
\begin{equation}
\E_{q_{\bphi}(\bx|\by)}\left\{x_{k}\right\} = \left( 2q_{\bphi,k}^{I}(\by) - 1 \right)
+ j \cdot \left( 2q_{\bphi,k}^{Q}(\by)-1 \right)
\label{eq:Exk}
\end{equation}
Hence, for the case where $k\neq l$ we have
\begin{align}
\MoveEqLeft[0.1]
\E_{q_{\bphi}(\bx|\by)}\left\{x_{k}\conj{x_{l}}\right\} = 
\E_{q_{\bphi}(\bx|\by)}\left\{x_{k}\right\} \cdot \E_{q_{\bphi}(\bx|\by)}\left\{\conj{x_{l}}\right\}
\label{eq:Exkxl}
\\
&=\left[ \left( 2q_{\bphi,k}^{I}(\by) - 1 \right) \left( 2q_{\bphi,l}^{I}(\by) - 1 \right) + 
\left( 2q_{\bphi,k}^{Q}(\by) - 1 \right) \left( 2q_{\bphi,l}^{Q}(\by) - 1 \right) \right] +
\\
&\quad j\cdot 
\left[ \left( 2q_{\bphi,k}^{Q}(\by) - 1 \right)
\left( 2q_{\bphi,l}^{I}(\by) - 1 \right) - \left( 2q_{\bphi,k}^{I}(\by) - 1 \right)
\left( 2q_{\bphi,l}^{Q}(\by) - 1 \right)\right]
\label{eq:Exkxl1}
\end{align}
We also have
\begin{equation}
\E_{q_{\bphi}(\bx|\by)}\left\{x_{k}\conj{x_{k}}\right\} = 2
\label{eq:Exkxk}
\end{equation}
Using \eqref{eq:Exk}, \eqref{eq:Exkxl1} and \eqref{eq:Exkxk} in \eqref{eq:C}, it is straight-forward to obtain an explicit expression for $C$. However, in order to compute the third term in the summation over $n$ efficiently, we use the fact that
\begin{align}
\MoveEqLeft[2]
\sum_{k,l=0}^{N-1}\E_{q_{\bphi}(\bx|\by)}\left\{x_{k}\conj{x_{l}}\right\}h_{n-k}\conj{h_{n-l}} 
=\\
&\left| \sum_{k=0}^{N-1} \E_{q_{\bphi}(\bx|\by)}\left\{x_{k}\right\}h_{n-k} \right|^2
+\\
&\sum_{k=0}^{N-1} \left| h_{n-k} \right|^{2}
\left[2- \left| \E_{q_{\bphi}(\bx|\by)}\left\{x_{k}\right\}\right|^{2} \right]
\label{eq:C1}
\end{align}
which follows from \eqref{eq:Exkxl} and \eqref{eq:Exkxk}. It is now straightforward to use \eqref{eq:Exk}, \eqref{eq:Exkxl1}, \eqref{eq:Exkxk} and \eqref{eq:C1} in \eqref{eq:C}, and obtain
\begin{align}
C = \sum_{n=0}^{N-1} \left[ \left| y_n \right|^2 - 2\alpha_n + \beta_n \right]
\label{eq:C2}
\end{align}
where
\begin{align}
\alpha_n
&=
\sum_{k=0}^{N-1} \left\{ y_{n}^{I}\cdot \left[h_{n-k}^{I} \left( 2q_{\bphi,k}^{I}(\by) - 1 \right) -
h_{n-k}^{Q} \left( 2q_{\bphi,k}^{Q}(\by) - 1 \right) \right] + \right.\\
&\qquad \left. y_{n}^{Q}\cdot \left[h_{n-k}^{Q} \left( 2q_{\bphi,k}^{I}(\by) - 1 \right) + 
h_{n-k}^{I} \left( 2q_{\bphi,k}^{Q}(\by) - 1 \right)\right] \right\}
\label{eq:alpha}
\end{align}
and
\begin{align}
\beta_n
&=
\left[ \sum_{k=0}^{N-1} h_{n-k}^{I} \left( 2q_{\bphi,k}^{I}(\by) - 1 \right) - 
h_{n-k}^{Q} \left( 2q_{\bphi,k}^{Q}(\by) - 1 \right) \right]^{2}\\
&+
\left[ \sum_{k=0}^{N-1} h_{n-k}^{Q} \left( 2q_{\bphi,k}^{I}(\by) - 1 \right) + 
h_{n-k}^{I} \left( 2q_{\bphi,k}^{Q}(\by) - 1 \right) \right]^{2}\\
&+
\sum_{k=0}^{N-1}\left[ \left( h_{n-k}^{I}\right)^{2}+\left( h_{n-k}^{Q}\right)^{2}\right] \cdot\\
&\qquad\quad
\left[4q_{\bphi,k}^{Q}(\by) + 4q_{\bphi,k}^{I}(\by) - 
4\left( q_{\bphi,k}^{I}(\by) \right)^{2} - 4\left( q_{\bphi,k}^{Q}(\by) \right)^{2} \right]
\label{eq:beta}
\end{align}

Now, to train our VAEE model, we need to minimize $\cL(\btheta,\bphi,\by) = -A-B$ with respect to $\btheta=\{\bh,\sigma_w^2\}$ and $\bphi$. We start with the minimization with respect to $\sigma_w^2$. Note that $A$ is independent of $\sigma_w^2$, and $B$ depends on $\sigma_w^2$ as described in \eqref{eq:B}. Hence, by setting the derivative of $B$ with respect to $\sigma_w^2$ to zero, we obtain that the optimal value of $\sigma_w^2$ is given by
$\sigma_w^2 = C / N$.
Using this and \eqref{eq:A} we see that up to an additive constant (which does not influence the gradients of the learned parameters $\btheta,\bphi$), the loss function $\cL(\bh,\bphi,\by)$ (using $\sigma_w^2=C/N$) for QPSK modulation is given by
\begin{equation}
\cL_{\text{QPSK}}(\bh,\bphi,\by) = \cL(\bh,\bphi,\by) = N \log C - \cH \left[ q_{\bphi}(\bx|\by) \right]
\label{eq:loss}
\end{equation}
where $\cH \left[ q_{\bphi}(\bx|\by) \right]$ is given in \eqref{eq:Hq}, and $C$ is given in \eqref{eq:C2}, \eqref{eq:alpha} and \eqref{eq:beta}.

The unknown parameters in our VAEE model are $\bh$ and $\bphi$. We estimate these parameters by applying gradient descent based optimization on the loss function $\cL(\bh,\bphi,\by)$ defined in \eqref{eq:loss}.
In our simulation program, the gradient of the loss function with respect to $\bh$ and $\bphi$ was calculated automatically by the Tensorflow framework \cite{abadi2016tensorflow}. It can be verified that the complexity (mainly multiplications and additions) of this computation scales linearly with respect to $N$. It also scales linearly with respect to the number of free parameters which include the $M$ coefficients of the channel impulse response and the convolution kernels of the neural network.

Recalling the definition of $C$ in \eqref{eq:B}, we see that our loss function, $\cL(\bh,\bphi,\by)$ defined in \eqref{eq:loss}, consists of the entropy of the variational approximation to the posterior, $\cH \left[ q_{\bphi}(\bx|\by) \right]$, that we wish to maximize (this is reasonable due to the i.i.d assumption of the symbols), and an autoencoder distortion term, $N\log C$, that we wish to minimize.

Our method provides an estimated channel response, $\bh$, as part of the learning process.
Note that we do not have to know the exact value of $M$. Instead, it suffices to have an upper bound on the order of the channel impulse response (or an upper bound on the order of a finite impulse response with which the true channel impulse response can be well approximated).
Also note that the VAE-encoder network outputs a soft decoding, $\{ q_{\bphi,j}^I(\by), q_{\bphi,j}^Q(\by) \}$, of the transmitted data.

\subsection{BPSK modulation} \label{sec:lin_chan_BPSK}
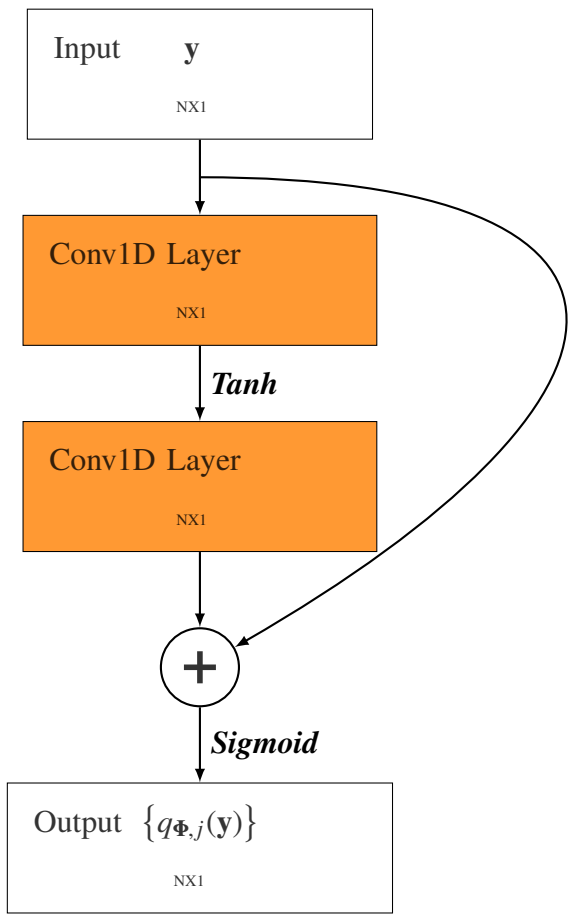
\begin{figure}[]
\centering
\scalebox{1}{\begin{tikzpicture}
\begin{scope}[start chain = going below,
every node/.append style={on chain,fill opacity=0.8,draw,
	join},every join/.style={thick,-latex},
cs/.style={minimum width=4.5cm,copy shadow={shadow scale=1, shadow xshift=0.5ex, shadow yshift=-0.5ex}}
]
\node[fill=white] (N1) {\begin{tabular}{p{1cm}cp{1cm}}
	Input &\enspace $\by$\quad\quad &\\
	& \enspace\tiny NX1\quad\quad \quad & 
	\end{tabular}};
\node[fill=orange] (N2) {\begin{tabular}{p{4cm}cp{1cm}}
	Conv1D Layer\\
	\qquad\qquad\tiny NX1 \qquad
	\end{tabular}};
\node[fill=orange] (N3) {\begin{tabular}{p{4cm}cp{1cm}}
	Conv1D Layer\\
	\qquad\qquad\tiny\tiny NX1 \qquad
	\end{tabular}};
\node[circle,minimum size=8mm,thick,font=\Huge\bfseries] (C) {+};
\node[fill=white] (N4) {\begin{tabular}{p{1cm}cp{1cm}}
	Output & $\left\{ q_{\bphi,j}(\by) \right\}$ &\\
	& \tiny NX1 \enspace\enspace& 
	\end{tabular}};
\end{scope}  
\path (N1) -- (N2) coordinate[pos=0.5] (aux);
\draw[thick,-latex] (aux) to[out=0,in=30,looseness=2.7] (C);
\path (N2) -- (N3) node[pos=0.5,right,font=\bfseries\itshape]{Tanh};
\path (C) -- (N4) node[pos=0.5,right,font=\bfseries\itshape]{Sigmoid};
\end{tikzpicture}}
\caption{Our equalizer's architecture for BPSK communication.}
\label{fig:architecture_VAD}
\end{figure}

The derivation above can be degenerated to BPSK modulation, where the transmitted symbols are $x_k \in \{1,-1\}$, representing the bits $c_k \in \{0,1\}$ (such that $x_k=(-1)^{c_k}$, and the noise is real and Gaussian with expectation zero and variance $\sigma_{w}^2$.
In this case, denoting
\begin{equation}
q_{\bphi,j}(\by) \defined q_{\bphi,j}(X_{j}=1|\by)
\label{eq:q_phi_j_y}
\end{equation}
we have
\begin{equation}
q_{\bphi}(\bx\given\by)
=
\prod_{j=0}^{N-1} 
\left( q_{\bphi,j}(\by) \right)^{(1+x_j)/2} \left( 1-q_{\bphi,j}(\by) \right)^{(1-x_j)/2}
\end{equation}
The loss is
\begin{equation}
\cL_{\text{BPSK}}(\bh,\bphi,\by) = -A - B
\end{equation}
where
\begin{align}
A &= \sum_{\bx} q_{\bphi}(\bx|\by) \cdot \left(  \log p(\bx) - \log q_{\bphi}(\bx|\by) \right)\\
&= -N\log 2 + \cH \left[ q_{\bphi}(\bx|\by) \right]
\label{eq:A_BPSK}
\end{align}
and
\begin{align}
\mathcal{H}\left[ q_{\bphi}(\bx|\by) \right]
&=
\mathcal{H}\left[ \prod_{j=0}^{N-1}q_{\bphi,j}(x_{j}|\by) \right]
=\sum_{j=0}^{N-1}  \cH \left[ q_{\bphi,j}(x_{j}|\by) \right]
\\
&= -\sum_{j=0}^{N-1} \left[ q_{\bphi,j}(\by) \log q_{\bphi,j}(\by) + 
\left( 1 - q_{\bphi,j}(\by) \right) \log \left(1 - q_{\bphi,j}(\by) \right) \right]
\label{eq:Hq_BPSK}
\end{align}
Instead of the QPSK loss, Eq. \eqref{eq:loss}, we now have
\begin{equation}
\cL_{\text{BPSK}}(\bh,\bphi,\by) = \frac{N}{2} \log C - \cH \left[ q_{\bphi}(\bx|\by) \right]
\label{eq:lossBPSK}
\end{equation}
for
\begin{align}
C &\defined \E_{q_{\bphi}(\bx|\by)} \left[ \left\lVert \by-\bx*\bh \right\rVert^{2} \right] 
\\
&= \sum_{n=0}^{N-1} \left\{ \left(y_{n}\right)^{2} - 2y_{n}
\sum_{k=0}^{N-1} \left( 2q_{\bphi,k}(\by) - 1 \right) h_{n-k} \right.
\\
&\qquad+ \left[ \sum_{k=0}^{N-1} \left( 2q_{\bphi,k}(\by) - 1 \right) h_{n-k} \right]^{2}\\
&\qquad+ \left. \sum_{k=0}^{N-1} \left( h_{n-k} \right)^{2} 
\left[ 4q_{\bphi,k}(\by) - 4\left( q_{\bphi,k}(\by) \right)^{2} \right] \right\}
\label{eq:C_BPSK}
\end{align}
Our VAE-encoder network for BPSK signaling, implementing $\{ q_{\bphi,j}(\by) \}$, is depicted in Fig. \ref{fig:architecture_VAD}. Unlike the QPSK case, we now use standard real one dimensional convolutional layers.

\section{Proposed method for a noisy nonlinear ISI channel} \label{sec:nonlin_chan}
Some authors have considered the nonlinear channel equalization problem, but only under a supervised learning setup, which requires pilot signals for training the equalizer, \cite{patra2009nonlinear, ye2017initial, 8491056, olmos2010joint, mitchinson2002digital}.
We now consider the extension of our unsupervised blind VAEE to noisy nonlinear ISI channels.
The channel model is described by \eqref{eq:sys} and \eqref{eq:conv} in Section \ref{sec:problem_setup} with an unknown, possibly nonlinear, function $g(\cdot)$ (also shown in Fig. \ref{fig:sys}) in addition to the other unknown channel parameters. Throughout this section we assume BPSK modulation, but the same derivation can be used for QPSK modulation.

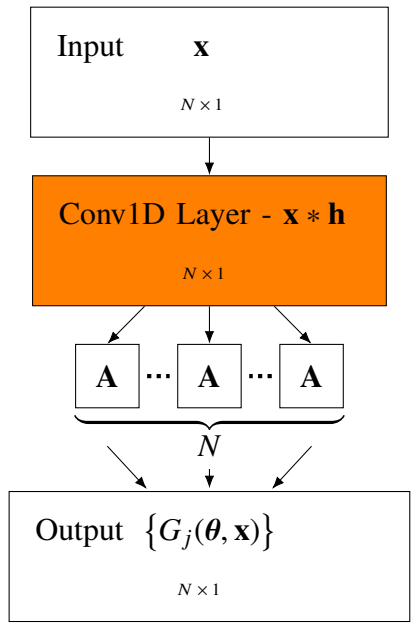
\begin{figure}[]
	\centering
	\scalebox{1}{\begin{tikzpicture}[
node distance = 5mm and 7mm,
start chain = going below,
box/.style = {draw, fill=#1, minimum size=2em},
box/.default = white,
every edge/.style = {draw, -Latex},
BC/.style = {decorate,
	decoration={calligraphic brace, amplitude=5pt,
		raise=1mm},
	very thick, pen colour=#1
}
]
\begin{scope}[every node/.append style={on chain}]
\node (N1) [box]
{\begin{tabular}{p{1cm}cp{1cm}}
	Input &\enspace $\bx$\quad\quad &\\
	& \enspace\tiny $N\times1$\quad\quad \quad & 
	\end{tabular}};
\node (N2) [box=orange]
{\begin{tabular}{p{4cm}cp{1cm}}
	Conv1D Layer - $\bx*\bh$\\
	\qquad\tiny\quad\quad\hspace{1.4em} $N\times1$ 
	\end{tabular}};
\node (N3) [box, node font=\bfseries] {A};
\end{scope}
\node (N4) [box,below=11mm of N3]
{\begin{tabular}{p{1cm}cp{1cm}}
	Output & $\left\{G_{j}(\boldsymbol{\theta},\bx)\right\}$ &\\
	& \tiny $N\times1$ \enspace\enspace& 
	\end{tabular}};
\node (N3L) [box, node font=\bfseries, left=5mm of N3] {A};
\node (N3R) [box, node font=\bfseries,right=5mm of N3] {A};
\path   (N1) edge (N2)
(N2) edge (N3)
(N2) edge (N3L.north)
(N2) edge (N3R.north);
\draw[BC]   (N3R.south east) -- node (aux) [below=2mm] {$N$} (N3L.south west);
\path[shorten >=1pt]    (aux -| N3L) edge (N4)
(aux)        edge (N4)
(aux -| N3R) edge (N4);
\draw[ultra thick, shorten <=1mm, shorten >=1mm, dotted]
(N3L) edge[-]  (N3)
(N3R) edge[-]  (N3);
\end{tikzpicture}}
	\caption{Our new VAE-decoder architecture, using a neural network A block (shown in Fig. \ref{fig:encoderA}).}
	\label{fig:encoder}
\end{figure}

Since the nonlinear function $g(\cdot)$ is unknown and needs to be learned, we use a VAE-decoder neural network in addition to the VAE-encoder neural network that we had in the linear channel case. Hence, for a noisy nonlinear ISI channel we are using a classical VAE structure, incorporating neural networks to model both the encoder and the decoder, thus using the full flexibility that the VAE approach offers. The VAE-decoder neural network has input $\bx$ and output $\bG(\bx,\btheta)$ where $\btheta$ is the neural network's parameter vector. The vector $\btheta$ includes both the unknown ISI channel response, $\bh$, and the neural network parameters, $\bpsi$, used to model the nonlinearity $g(\cdot)$, i.e., $\btheta=(\bh,\bpsi)$ . The VAE-decoder $\bG(\bx,\btheta)$ needs to implement (approximately) the function $\bg(\bx * \bh)$, i.e.,
\begin{align}
\bG\left( \bx, \btheta \right) \approx \bg\left(\bx * \bh \right)
\label{eq:func}
\end{align}
The VAE-decoder network we use, which acts as an equalizer in our problem to estimate the latent transmitted bit sequence, $\bx$, is described in Figs. \ref{fig:encoder}-\ref{fig:encoderA} and Table \ref{tab:fcn-layout}.
\begin{figure}[]
	\centering
	\scalebox{1}{\begin{tikzpicture}[cir/.style={circle,draw=#1,minimum size=0.5cm},y=0.6cm,font=\sffamily]
\begin{scope}[rotate=90]
\node[cir=blue!50!black] (a2) at (0,-2) {};
\node[cir=blue] (b1) at (2,0) {};
\node[cir=blue] (b2) at (2,-1) {};
\node[cir=blue] (b3) at (2,-2) {};
\node[cir=blue] (b4) at (2,-3) {};
\node[cir=blue] (b5) at (2,-4) {};
\node[cir=blue!50] (c1) at (4,0) {};
\node[cir=blue!50] (c2) at (4,-1) {};
\node[cir=blue!50] (c3) at (4,-2) {};
\node[cir=blue!50] (c4) at (4,-3) {};
\node[cir=blue!50] (c5) at (4,-4) {};
\node[cir=blue] (d1) at (6,0) {};
\node[cir=blue] (d2) at (6,-1) {};
\node[cir=blue] (d3) at (6,-2) {};
\node[cir=blue] (d4) at (6,-3) {};
\node[cir=blue] (d5) at (6,-4) {};
\node[cir=blue!50] (e1) at (8,0) {};
\node[cir=blue!50] (e2) at (8,-1) {};
\node[cir=blue!50] (e3) at (8,-2) {};
\node[cir=blue!50] (e4) at (8,-3) {};
\node[cir=blue!50] (e5) at (8,-4) {};
\node[cir=green!50!black] (x) at (10,-2) {};
\foreach \i/\j in {c/d} {
    \foreach \cnto in {1,2,3,4,5} {
        \foreach \cntt in {1,2,3,4,5} {
            \draw[-latex] (\j\cnto.south)--(\i\cntt.north);
        }
    }
}
\foreach \i/\j in {b/c,d/e} {
    \foreach \cnt in {1,2,3,4,5} \draw[-latex] (\j\cnt)--(\i\cnt);
}
\foreach \i in {1,2,3,4,5} \draw[-latex] (x.south)--(e\i.north);
\foreach \i in {1,2,3,4,5} \draw[-latex] (b\i.south)--(a2.north);
\end{scope}
\draw (b1) node[blue,left=1em] {Dropout};
\draw (c1) node[blue!50,left=1em] {ReLU};
\draw (d1) node[blue,left=1em] {Dropout};
\draw (e1) node[blue!50,left=1em] {ReLU};
\end{tikzpicture}}
	\caption{The neural network A in Fig. \ref{fig:encoder}.}
	\label{fig:encoderA}
\end{figure}
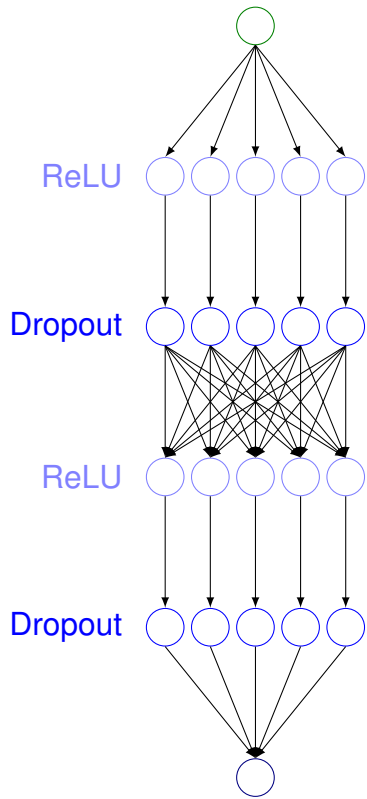
\begin{table}[htbp]
	\renewcommand{\arraystretch}{1.0} 
	\centering
	\caption{The VAE-decoder's Fully Connected Neural Network A Layout}
	\label{tab:fcn-layout}
	\begin{tabular}{l|l}
		Layer    & Output dimensions    \\
		\hline
		Input & $1$ \\
		FC/ReLU & $5$ \\
		Dropout & $5$ \\
		FC/ReLU & $5$ \\
		Dropout & $5$ \\
		Linear & $1$
	\end{tabular}
\end{table}
As can be seen in Fig. \ref{fig:encoder}, the VAE-decoder network first applies the convolution with $\bh$ using a convolutional layer. We then implement the function $g(\cdot)$ using the neural network A, shown in Fig. \ref{fig:encoderA}. The neural network A is applied component-wise on the results of the convolution with $\bh$. It uses a fully connected (FC) architecture as described in Fig. \ref{fig:encoderA} and Table \ref{tab:fcn-layout}, including ReLU activation functions, dropout layers \cite{srivastava2014dropout} and a linear output layer.


Repeating the derivation in Section \ref{sec:lin_chan_BPSK} for the nonlinear case, the new loss function $\cL_{\text{BPSK}}^{\text{NL}}(\btheta,\bphi,\by)$ is given by the same expression \eqref{eq:lossBPSK}, i.e.,
\begin{equation}
\cL_{\text{BPSK}}^{\text{NL}}(\btheta,\bphi,\by) = \frac{N}{2} \log C_{\text{NL}} - \cH \left[ q_{\bphi}(\bx|\by) \right]
\label{eq:lossNL}
\end{equation}
where $\mathcal{H}\left[ q_{\bphi}(\bx|\by) \right]$ is given by \eqref{eq:Hq_BPSK} for $q_{\bphi,j}(\by)$ defined by \eqref{eq:q_phi_j_y}. However, unlike the analytic computation of $C$ in Section \ref{sec:lin_chan} for linear channels, $C_{\text{NL}}$, which is defined by
\begin{align}
C_{\text{NL}}
&\defined \E_{q_{\bphi}(\bx|\by)} \left[ \left\lVert \by - \bg(\bx*\bh) \right\rVert^{2} \right]\\
&\approx \E_{q_{\bphi}(\bx|\by)} \left[ \left\lVert \by - \bG(\bx,\btheta) \right\rVert^{2} \right]\\
&= \E_{q_{\bphi}(\bx|\by)} \left\{ \sum_{n=0}^{N-1}\left[ y_n - G_n\left( \bx,\btheta \right) \right]^2 \right\}
\label{eq:C_NL}
\end{align}
($G_n(\cdot)$ is the $n$-th output cell of the neural network) cannot be computed analytically.
Our VAE-encoder network, implementing $q_{\bphi,j}(\by)$, is the same as in Section \ref{sec:lin_chan_BPSK} (Fig. \ref{fig:architecture_VAD}). Hence, we now have two neural networks, a VAE-decoder neural network implementing $\bG(\bx,\btheta)$, and a VAE-encoder neural network implementing $q_\bphi(\bx|\by)$. The two neural networks are trained jointly using a gradient descent approach. Hence, we need to compute the gradient of $\cL_{\text{BPSK}}^{\text{NL}}(\btheta,\bphi,\by)$ with respect to $\btheta$ and $\bphi$. By \eqref{eq:lossNL} and \eqref{eq:C_NL} we have
\begin{align}
\nabla_{\btheta} \cL_{\text{BPSK}}^{\text{NL}}(\btheta,\bphi,\by)
&= \frac{N}{2} \nabla_{\btheta} \log \E_{q_{\bphi}(\bx|\by)} \left\{ \sum_{n=0}^{N-1}\left[ y_n - G_n\left( \bx,\btheta \right) \right]^2 \right\} \label{eq:nabla_theta_L}\\
\nabla_{\bphi} \cL_{\text{BPSK}}^{\text{NL}}(\btheta,\bphi,\by)
&= \frac{N}{2} \nabla_{\bphi} \log \E_{q_{\bphi}(\bx|\by)} \left\{ \sum_{n=0}^{N-1}\left[ y_n - G_n\left( \bx,\btheta \right) \right]^2 \right\} - \nabla_{\bphi} \cH \left[ q_{\bphi}(\bx|\by) \right]
\label{eq:nabla_phi_L}
\end{align} 
The second term in the right hand side (RHS) of \eqref{eq:nabla_phi_L} can be easily computed from \eqref{eq:Hq_BPSK} (the derivative of $q_{\bphi,j}(\by)$ with respect to $\bphi$ is obtained by the backpropagation algorithm applied to the VAE-encoder neural network). Following the common approach \cite{kingma2013auto, nvil}, the gradient with respect to $\btheta$ in \eqref{eq:nabla_theta_L} can be well approximated using
\begin{equation}
\nabla_{\btheta} \cL_{\text{BPSK}}^{\text{NL}}(\btheta,\bphi,\by)
\approx
\frac{N}{2} \nabla_{\btheta} \log \left\{ \sum_{n=0}^{N-1}\left[ y_n - G_n\left( \hat{\bx},\btheta \right) \right]^2 \right\} \label{eq:nabla_theta_L_approx}
\end{equation}
where $\hat{\bx}$ is obtained by sampling using the Bernoulli distribution with probabilities given by the output of the VAE-encoder neural network that implements $\{ q_{\bphi,j}(\by) \}$, i.e., we set $\hat{x_j}=1$ ($\hat{x_j} = -1$, respectively) with probability $q_{\bphi,j}(\by)$ ($1-q_{\bphi,j}(\by)$).

However, it is more difficult to obtain a reliable estimate to the first term in the RHS of \eqref{eq:nabla_phi_L}.
When $\bx$ in the model is continuous (e.g., a Gaussian random variable), the reparameterization trick can be used \cite{kingma2013auto}. For discrete $\bx$ (as in our problem), the reparametrization trick cannot be applied.
Instead, various approximation schemes to the gradient have been suggested, e.g., \cite{nvil,vimco}. 
In \cite{maddison2016concrete,jang2017categorical} the gradient is approximated using continuous relaxations of discrete distributions.
This approximation was shown to possess a favorable trade-off between estimation quality, computational complexity and sample efficiency. Hence it was adopted in our work. 

The method in \cite{maddison2016concrete,jang2017categorical} is based on a continuous relaxation of the Gumbel-Max trick \cite{gumbel1954,luce1959}, which allows us to sample from a categorical distribution (in our case this is the Bernoulli distribution with probabilities $(q_{\bphi,j}(\by),1-q_{\bphi,j}(\by))$).
Denote by $\text{U}(0,1)$ the standard uniform distribution. We repeat the following procedure for $j=0,1,\ldots,N-1$: We first sample $u_{j,k} \sim \text{U}(0,1)$, and set $g_{j,k} = -\log\left(-\log u_{j,k}\right)$ for $k=1,2$. By definition, $g_{j,1}$ and $g_{j,2}$ are said to be $\text{Gumbel}(0,1)$ distributed.
We now define, for $j=0,1,\ldots,N-1$,
\begin{align}
\hat{c}_j &\defined
\frac{\exp\left( \left( \log q_{\bphi,j}(\by) + g_{j,1} \right) / \tau \right)}
{\exp\left(\left( \log q_{\bphi,j}(\by) + g_{j,1} \right) / \tau \right) +
\exp\left(\left( \log\left( 1-q_{\bphi,j}(\by) \right) + g_{j,2} \right) / \tau \right)}\\
\hat{x}_j &\defined 2\hat{c}_j - 1
\label{eq:softmax_app}
\end{align}
where $\tau>0$ is some parameter (temperature).
It can be shown \cite{maddison2016concrete,jang2017categorical}, that for $\tau\rightarrow 0$, the resulting $\hat{\bx}$ is a sample from the distribution $q_{\bphi}(\bx|\by)$ ($\hat{c}_j$ is the sampled soft bit value, and $\hat{x}_j$ is the corresponding sampled soft BPSK modulated symbol).
Our estimate to the first term in the RHS of \eqref{eq:nabla_phi_L} is then
\begin{equation}
\frac{N}{2} \nabla_{\bphi} \log \sum_{n=0}^{N-1}\left[ y_n - G_n\left( \hat{\bx},\btheta \right) \right]^2
\end{equation}
We set $\tau>0$ to keep $\hat{x}_j$, defined in \eqref{eq:softmax_app}, smooth and differentiable with respect to $\bphi$.
In our simulations we initialize the temperature to $\tau=5$ and set it to be trainable as recommended in \cite{maddison2016concrete,jang2017categorical}.

We note that our stochastic approximation \eqref{eq:nabla_theta_L_approx} to the gradient \eqref{eq:nabla_theta_L} is biased. This is due to the fact that by optimizing the loss analytically with respect to $\sigma_w$ (recall Section \ref{sec:lin_chan_QPSK}) we obtained a log-expectation expression in \eqref{eq:nabla_theta_L}. As an alternative to the analytic optimization of $\sigma_w$, one could estimate $\sigma_w$ using stochastic gradient descent. The advantage of doing so is that the stochastic approximation to the gradient of $\btheta$ will then be unbiased. The drawback is that we would then not use the fact that $\sigma_w$ can be optimized analytically. A similar note applies to the stochastic approximation to the gradient \eqref{eq:nabla_phi_L} with the following difference: In this case, the continuous relaxation of the Gumbel-Max trick already creates a bias in the gradient estimation. A thorough evaluation of this topic is left to future research.

\section{Extension to LDPC coded communication} \label{coded_data}
In order to enable reliable communications at rates close to channel capacity, an error correcting code needs to be incorporated. In this section we assume the availability of only the channel observations corresponding to a single transmitted codeword, without knowing the channel parameters and without using pilot signals that reduce the communication rate. The goal is to reconstruct the transmitted message from this data alone in an unsupervised way. It is useful for a fast changing communication (e.g., wireless) environment.

Throughout the section we assume a noisy linear ISI channel and BPSK modulation. However, the same derivation can be applied to a noisy nonlinear ISI channel by incorporating our method in Section \ref{sec:nonlin_chan}.
In the simulations section we report results for both the linear and nonlinear cases under BPSK modulation.
The results can also be easily extended to other modulation schemes (e.g., QPSK). 
We discuss the case where the transmitted data, $\bx$, is a BPSK modulated codeword, $\bc$ (i.e. $x_i=(-1)^{c_i}$), of a sparse graph-based code. For concreteness, in this paper we assume an LDPC code \cite{gallager_LDPC_article}.
However, our methods can be used for other linear sparse graphical codes such as turbo codes, to which BP decoding can be applied.

A binary LDPC code, $\cC$, is a binary linear code that can be described by a sparse binary parity check matrix $H$ of dimensions $J \times N$, such that $\cC = \{ \bc \: : \: H\bc = {\bf 0}  \}$. The blocklength of the code is $N$, and the code rate is at least $(N-J)/N$ (due to a possible linear dependence between the rows of $H$).
The matrix $H$ can also be represented by a \emph{Tanner graph}, $\cG$, \cite{ru_book} which is a sparse bipartite graph, with $N$ left nodes, $\cI$, also called variable nodes, and $J$ right nodes, $\cJ$, also called parity check nodes. A variable node $i\in\cI$ (parity check node $j\in\cJ$, respectively), can only connect to parity check (variable) nodes. We denote this set of neighbor nodes by $\cN_i$ ($\cN_j$, respectively). An edge connects the parity check node $j\in\cJ$ and the variable node $i\in \cI$ if and only if $H_{j,i}=1$. LDPC codes can be efficiently decoded using Gallager's probabilistic decoding algorithm, also known as the sum-product or BP algorithm. This algorithm is a message passing algorithm over edges in the Tanner graph \cite{ru_book}.

We describe two methods for enhancing the operation of our VAEE when $\bx$ is a BPSK modulated LDPC codeword.
In section \ref{sec:VAE_Gal_lemma} we add a loss term that penalizes the soft decoding based on the estimated probabilities that the check nodes in the Tanner graph are not satisfied.
In section \ref{sec:VAE_Turbo} we suggest a decoding scheme that applies the VAEE followed by the BP decoding algorithm iteratively similarly to \cite{liang2018iterative}.
In our experiments we observed that both methods were useful to improve decoding, and the best results were obtained by using both simultaneously (results are provided below in the simulations section).

\subsection{Augmenting the loss using Gallager's lemma} \label{sec:VAE_Gal_lemma}
In \cite[Lemma 1]{gallager_LDPC_article}, Gallager proved the following. Consider $m$ statistically independent bits, where the $i$'th bit is 1 with probability $P_i$ and 0 with probability $1-P_i$. Then the probability that an even number of bits are 1 is
\begin{equation}
\frac{1+\prod_{i=1}^m \left( 1 - 2 P_i \right)}{2}
\label{eq:gal_lemma}
\end{equation}
Recall that $q_{\bphi,i}(\by)$ ($1-q_{\bphi,i}(\by)$, respectively) is an estimate to the probability that $x_i=1$ ($x_i=-1$), corresponding to $c_i=0$ ($c_i=1$). Hence, by \cite[Lemma 1]{gallager_LDPC_article}, for any parity check node, $j\in\cJ$, the probability that the check node is satisfied, i.e., an even number of variable nodes $i\in\cN_j$ satisfy $c_i=1$, can be estimated by (we set $P_i=1-q_{\bphi,i}(\by)$ in \eqref{eq:gal_lemma} since this is the probability that $c_i=1$, corresponding to $x_i=-1$)
\begin{equation}
\frac{ 1 + \prod_{i\in \cN_j} \left( 2 q_{\bphi,i}(\by) - 1 \right)}{2}
\end{equation}
Now, for a valid codeword $\bc\in \cC$, all check nodes are satisfied. Hence we request a low value to the following \emph{Gallager loss} defined by,
\begin{align}
\cL_{\text{G}}(\bphi,\by) &= -\log \prod_{j\in \cJ} \frac{ 1 + \prod_{i\in \cN_j} \left( 2 q_{\bphi,i}(\by) - 1 \right)}{2}
\\
&= -\sum_{j\in \cJ}\log \frac{ 1 + \prod_{i\in \cN_j} \left( 2 q_{\bphi,i}(\by) - 1 \right)}{2}
\label{eq:lossGallager}
\end{align}

Instead of $\cL_{\text{BPSK}}(\bh,\bphi,\by)$ in \eqref{eq:lossBPSK}, we thus propose the following augmented loss for the coded data case
\begin{align}
\cL_{\text{BPSK}}^{\text{G}}(\bh,\bphi,\by) &= \lambda \cdot \cL_{\text{BPSK}}(\bh,\bphi,\by) + 
(1-\lambda) \cdot \cL_{\text{G}}(\bphi,\by)
\label{eq:lossVAD}
\end{align}
where $\lambda\in \left[0,1 \right]$ is a hyper-parameter determining how much weight is assigned to each component of the total loss.
Note that for $\lambda=0$ our loss function is just $\cL_{\text{G}}(\bphi,\by)$, and the VAE-encoder can produce the trivial solution, $q_{\bphi,i}(\by) = 1$ for all $i$, corresponding to the zero codeword. Hence, we must set $\lambda>0$.

Our Gallager loss, $\cL_{\text{G}}(\bphi,\by)$, is similar to the \emph{syndrome loss} introduced in \cite{lugosch2018learning}. However, in \cite{lugosch2018learning} the syndrome loss is used to improve the training of a neural message passing decoder, while in our work the Gallager loss is used for blind channel equalization of coded data. In addition, our loss is a likelihood based score.

\subsection{Iterative VAE Equalization and BP decoding} \label{Using_Turbo_mode} \label{sec:VAE_Turbo}
We now extend the VAEE to a turbo VAEE algorithm that applies VAEE and BP decoding iteratively, similarly to the turbo equalization algorithm \cite{douillard1995iterative,koetter2004turbo}.
We start the first iteration of the turbo VAEE algorithm by applying the VAEE. The prior probability of the transmitted binary data, $\bx$ (where $x_i\in\{1,-1\}$), is then uniform
\begin{equation}
p(\bx) = p^{(1)}(\bx) = \prod_i p^{(1)}_i(x_i)= 2^{-N}
\label{eq:p0}
\end{equation}
as in \eqref{eq:px_BPSK}.
The output of the VAEE (with or without the Gallager loss term that incorporates some coding information) are the probabilities $\{ q_{\bphi,i}(\by) = q_{\bphi,i}(X_i=1 | \by) \}$. We then apply the BP algorithm using these probabilities, produced by the VAEE, as uncoded data from the channel. The corresponding uncoded log-likelihood ratios (LLRs) at the input to the BP algorithm are
\begin{equation}
\text{LLR}^{(1)}_{i} = \log \frac{q_{\bphi}(X_{i}=1\given\by)}{q_{\bphi}(X_{i}=-1\given\by)}= \log \frac{q_{\bphi,i}(\by)}{1-q_{\bphi,i}(\by)} 
\label{eq:llr1}
\end{equation}
for $i=0,1,\ldots,N-1$. The outputs of the BP are soft decoding LLRs of the transmitted codeword, denoted by
\begin{equation}
\overline{\text{LLR}}^{(1)}_{i} = \log \frac{\overline{p}^{(1)}_i(X_i=1)}{\overline{p}^{(1)}_i(X_i=-1)}
\label{eq:llr2}
\end{equation}
where $\overline{p}^{(1)}_i(X_i=x_i)=\overline{p}^{(1)}_i(x_i)$ are the probabilities obtained by the BP algorithm.
Note that according to the principles of message passing algorithms \cite{ru_book}, for each $i$ the final marginalization used to obtain $\overline{\text{LLR}}^{(1)}_{i}$ does not include the input channel LLR message $\text{LLR}_i^{(1)}$. By \eqref{eq:llr2}, and since $\overline{p}^{(1)}_i(X_i=-1)=1-\overline{p}^{(1)}_i(X_i=1)$, $\overline{p}^{(1)}_i(X_i=1)$ can be extracted from $\overline{\text{LLR}}^{(1)}_{i}$ using
\begin{equation}
\overline{p}^{(1)}_i(X_i=1) = \left(1+e^{-\overline{\text{LLR}}^{(1)}_{i}}\right)^{-1}
\label{eq:llr2p}
\end{equation}
We now move on to the second turbo VAEE iteration by applying VAEE using
\begin{equation}
p(\bx) = p^{(2)}(\bx) = \prod_i \overline{p}^{(1)}_i(x_i)
\label{eq:p1}
\end{equation}
as prior probabilities of the transmitted data, $\bx$.
Recall that the VAEE was derived under the assumption of uniform $p(\bx)$ as in \eqref{eq:px_BPSK} and \eqref{eq:p0}. Hence, we need to generalize the algorithm to the case where the prior probability is given in \eqref{eq:p1}. This is easy, however, since the only change in the VAEE training is in the computation of the term $A$, which was previously calculated using \eqref{eq:A_BPSK}, and is now computed as follows,
\begin{align}
A &= \sum_{\bx} q_{\bphi}(\bx|\by) \left(  \log p(\bx) - \log q_{\bphi}(\bx|\by) \right)\\
&= \sum_{\bx} q_{\bphi}(\bx|\by) \sum_i \log p_i(x_i) + \cH \left[ q_{\bphi}(\bx|\by) \right]\\
&= \sum_{\bx} \prod_{l=0}^{N-1} q_{\bphi}(x_{l}|\by) \sum_i \log p_i(x_i) + \cH \left[ q_{\bphi}(\bx|\by) \right]\\
&= \sum_i \sum_{x_i} q_{\bphi}(x_{i}|\by) \log p_i(x_i) + \cH \left[ q_{\bphi}(\bx|\by) \right]\\
&= \sum_i \sum_{x \in \left\{-1,1\right\}} q_{\bphi,i}(x|\by) \log p_i(x) + \cH \left[ q_{\bphi}(\bx|\by) \right]
\label{eq:A1}
\end{align}
The corrected loss to be minimized instead of \eqref{eq:lossBPSK} is thus
\begin{equation}
\cL'_{\text{BPSK}}(\bh,\bphi,\by) = \frac{N}{2} \log C - \cH \left[ q_{\bphi}(\bx|\by) \right]
-\sum_i \sum_{x \in \left\{-1,1\right\}} q_{\bphi,i}(x|\by) \log p_i(x)
\label{eq:lossBPSK_Turbo}
\end{equation}
where $C$ is given in \eqref{eq:C_BPSK} (the Gallager loss $\cL_{\text{G}}$ can also be used by adding it to the above loss term).

The same procedure is repeated in the other decoding iterations. In general, in the first stage of the $l$-th iteration ($l=1,2,\ldots$), our proposed turbo VAEE applies the VAEE, by minimizing \eqref{eq:lossBPSK_Turbo} (possibly with the addition of the Gallager loss, $\cL_G$, as in \eqref{eq:lossVAD}) using
\begin{equation}
p_i(x) = p^{(l)}_i(x) = \overline{p}^{(l-1)}_i(x)
\label{eq:pjx}
\end{equation}
for $i=0,1,\ldots,N-1$ and $x\in\{1,-1\}$ as input (for initialization, $l=1$, we use $p_i(x)=p_i^{(1)}(x)=1/2$ as was described above). The VAEE produces the probabilities $q_{\bphi,i}(\by) = q_{\bphi,i}(X_i=1 | \by)$. Then, in the second stage of the $l$-th iteration, we apply the BP algorithm, using $q_{\bphi,i}(\by)$ as uncoded probabilities from the channel. The BP algorithm produces the probabilities $\overline{p}_i^{(l)}(x)$, $x\in\{1,-1\}$ as in the first iteration: $\overline{\text{LLR}}^{(l)}_{i}$ is obtained from variable node final marginalization, by summing all incoming final LLR messages to variable node $i$ in the Tanner graph, except for the input channel LLR message, $\text{LLR}_i^{(l)}=\log \frac{q_{\bphi,i}(\by)}{1-q_{\bphi,i}(\by)}$, and then $\overline{p}^{(l)}_i(X_i=1) = \left(1+e^{-\overline{\text{LLR}}^{(l)}_{i}}\right)^{-1}$ as in \eqref{eq:llr2p}.
These probabilities are subsequently used as input to the next ($l+1$) iteration of the VAEE.

Due to the presence of short cycles in the code Tanner graph, 
the (loopy) BP algorithm is not accurate. It tends to overestimate the reliabilities of the estimated bits. Hence,
we found it useful to weaken (pull back towards $1/2$) the prior probabilities used in the beginning of each iteration by modifying \eqref{eq:pjx} to 
\begin{equation}
p_i(x) = p^{(l)}_i(x) = \alpha \overline{p}^{(l-1)}_i(x) + (1-\alpha) \frac{1}{2}
\label{eq:pjx_1}
\end{equation}
for $i=0,1,\ldots,N-1$ and $x\in\{1,-1\}$, where $\alpha$ is some hyper-parameter.
Similarly, we found it useful to weaken the LLRs passed from the VAEE to the BP decoder by attenuating these by a factor of $\eta\in(0,1)$.

\begin{figure}[]
\centering
\scalebox{0.75}{\tikzset{%
	sum/.style      = {draw, circle, node distance = 2cm}, 
	input/.style    = {coordinate}, 
	output/.style   = {coordinate}, 
	block/.style = { draw,
		thick,
		rectangle,
		minimum height = 2em,
		fill=white,
		align=center,
		minimum width = 2em
	},
	wide block/.style = {
		block,
		minimum height = 3em,
		text width=2.2cm,
		minimum width = 2.5em
	},
	dotted_block/.style={draw=black!20!white, line width=1pt, dash pattern=on 1pt off 4pt on 6pt off 4pt,
		inner ysep=6mm,inner xsep=1mm, rectangle, rounded corners},
	virtual/.style = {coordinate},
	adder/.style={draw,circle}

}
\newcommand{\suma}{\Large$+$}

\begin{tikzpicture}[auto, thick, node distance=2cm, >=triangle 45]
\draw
node at (0,0){}
node [input, name=input1] {} 
node [align=center, wide block, right = 1cm of input1] (inte2) {LDPC Encoder}
node [align=center, wide block, right = 1cm of inte2] (inte3) {BPSK modulator}
node [align=center, wide block, right = 1cm of inte3] (inte4) {Intersymbol intereference}
node [sum, right = 1cm of inte4] (suma1) {\suma}
node [input, name=input2, above = 1cm of suma1] {} 
node [output, name=output1, right = 1cm of suma1] {};
\node [align=center, wide block, right = 1.3cm of output1] (glamor) {Variational Autoencoder Equalizer};
\node[wide block, right = 16mm of glamor] (trainer) {Belief Propagation};
\node[block, below=10mm of trainer](L){$D$};
\node[block, left=24mm of L](M){$\alpha$};
\node[sum, left=24mm of M](A1){+};
\coordinate[below=28mm of inte2](M1){};
\coordinate[below=28mm of inte3](L1){};
\coordinate[below=26mm of inte4](L2){};
\coordinate[below=4mm of trainer](M3){};
\coordinate[below=4mm of glamor](L3){};
\draw[->](input1) -- node (kmm1) {$\bm$}(inte2);
\draw[->](inte2) -- node {$\bc$}(inte3);
\draw[->](inte3) -- node (dx) {$\bx$}(inte4);
\draw[->](inte4) -- node {$\bx*\bh$}(suma1);
\draw[->](input2) -- node {$\bw$} (suma1);
\draw[<-](glamor.west) --node[above](YY){$\by$} ++(-2.3,0);
\draw[->](glamor) -- node {$\left\{q_{\Phi ,i}\left(\by\right)\right\}$} (trainer);
\draw[->](trainer.east) -- node[name=y]{$\left\{\overline{p}^{(l)}_i(x_i)\right\}$} ++ (2,0);
\draw[->,rounded corners](trainer.east) -- ++(1,0) |- (L);
\draw[->](L)--(M);
\draw[<-, rounded corners]([yshift=1mm]glamor.south west) 
-- ++(-1.92,0) node[above right] (secInput) {$\left\{p^{(l)}_{i}(x_i)\right\}$}   |- (A1);
\draw[<-] (A1) -- (M);
\draw[<-] (A1.south) -- ++ (0,-8mm) node[pos=0.7,right](BB){$\left(1-\alpha\right)\frac{1}{2}$};
\node [dotted_block, fit =(kmm1.west) (inte2) (inte3) (dx.west) (M1) (L1)] (aa) {};
\node [dotted_block, fit = (inte4) (suma1) (dx.east) (L2)] (aa2) {};
\node [dotted_block, fit = (glamor) (trainer) (M3) (L3) (y) (A1) (BB) (YY) (secInput)] (aa3) {};
\node at (aa.north) [above, inner sep=3mm] {Transmitter};
\node at (aa2.north) [above, inner sep=3mm] {Channel};
\node at (aa3.north) [above, inner sep=3mm] {Turbo VAEE};

\end{tikzpicture}}
\caption{Our proposed turbo VAEE which consists of iterative VAEE and BP decoding. $D$ is a delay element.}
\label{fig:turbo_arch}
\end{figure}
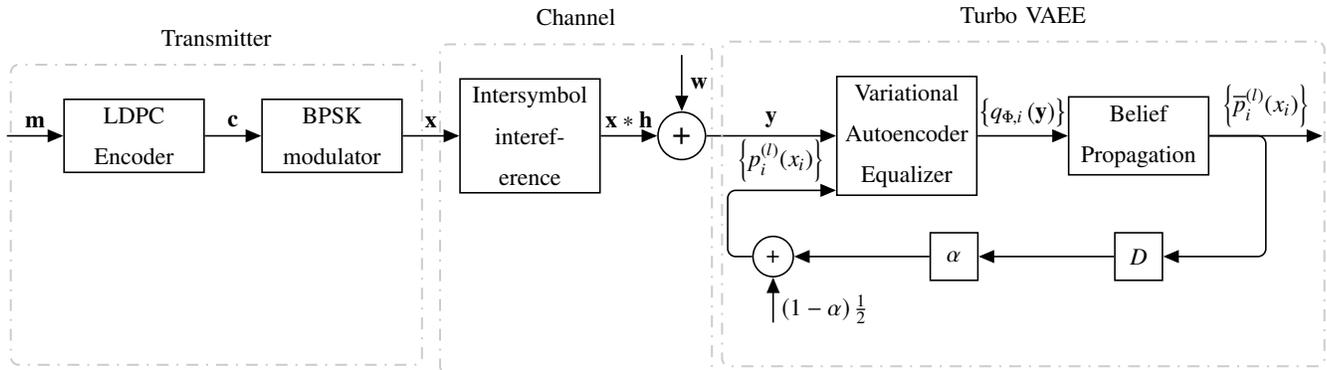

A high-level system scheme is shown in Fig. \ref{fig:turbo_arch}.
\rem{
Algorithm \ref{alg:TurboVAEE} summarizes the turbo VAEE algorithm. 
\newtoks\rowvectoks
\newcommand{\rowvec}[2]{%
	\rowvectoks={#2}\count255=#1\relax
	\advance\count255 by -1
	\rowvecnexta}
\newcommand{\rowvecnexta}{%
	\ifnum\count255>0
	\expandafter\rowvecnextb
	\else
	\begin{pmatrix}\the\rowvectoks\end{pmatrix}
	\fi}
\newcommand\rowvecnextb[1]{%
	\rowvectoks=\expandafter{\the\rowvectoks&#1}%
	\advance\count255 by -1
	\rowvecnexta}
\begin{algorithm}
Initialize $p^{(l=0)}(\bx)\gets(\frac{1}{2},\frac{1}{2},\ldots,\frac{1}{2})$;
	
\For{$t=1,\ldots,T$}{
	Train VAD for 1 step of with $p^{(l)}(\bx)$\;
	
	\If {$t \geq I$}{
		\For{$b=1,\ldots, B$}{
			Run BP algorithm over the VAD outputs\;
		}
		Subtract intrinsic information\;
		\ForEach {$j=1,2,\ldots,N;\: x \in \{0,1\}$}{
			\lefteqn{p_j(x) \gets p^{(l+1)}_j(x) = \alpha \overline{p}^{(l)}_j(x) + (1-\alpha) \frac{1}{2}}
		}
		$l\gets l+1$\;
	}
}
\caption{Turbo VAEE for unsupervised channel equalization and decoding}
\label{alg:TurboVAEE}
\end{algorithm}
}

\section{Simulation results} \label{sec:sim}
In our derivation above we assumed that both the channel impulse response, $\bh$, and the input signal $\{x_k\}$ are causal. Hence, considering for example the BPSK case, Eq. \eqref{eq:C_BPSK} (and similarly for the QPSK case), we sum over $k=0,1,\ldots,N-1$, and we do not need to consider negative values of $k$. However, if we are considering a sampled block $\by=(y_0,y_1,\ldots,y_{N-1})$ of $N$ measurements of the signal starting at some arbitrary time, then the above causality assumption on $\bx$ does not hold. Nevertheless, the edge effect decays as $N$ increases.
The causality assumption is equivalent to $M-1$ zero-padding of $\bx=(x_0,x_1,\ldots,x_{N-1})$ on the left. Alternatively (supposing odd $M$ for simplicity), we can assume that $\bh=(h_{-(M-1)/2},\ldots,h_0,\ldots,h_{(M-1)/2})$. Accordingly, we assume zero-padding of $\bx=(x_0,x_1,\ldots,x_{N-1})$ by $(M-1)/2$ both on the left and on the right, and the given measurements vector $\by=(y_0,y_1,\ldots,y_{N-1})$ is the result of the channel model \eqref{eq:conv}. As a result of this assumption we can still use the same equation \eqref{eq:C_BPSK} with summation over the same range of $k$. We used this second approach in our experiments with uncoded data, although the performance was similar to the performance of the first approach.

For the experiments with coded data we assumed that the transmitted BPSK modulated codeword $\bx=(x_0,x_1,\ldots,x_{N-1})$ starts after a random (and unknown) sequence of BPSK modulated bits. 
The filter $\bh=(h_0,h_1,\ldots,h_{M-1})$ is a causal size $M$ impulse response. 
The channel measurements are $\by=(y_0,y_1,\ldots,y_{N-1})$ ($N$ measurements).
The same conditions were used for the baseline methods that we compared with.

We implemented our VAEE algorithms using the Tensorflow framework \cite{abadi2016tensorflow} which provides automatic differentiation of the loss function. For the LDPC infrastructure, we used the software toolbox in \cite{channelcodes}. For the turbo operation mode, we used the BP algorithm implemented in \cite{liang2018iterative}.

\subsection{Linear channels, uncoded data} \label{sec:sim_lin_uncod}

\begin{figure}
	\centering
	\includegraphics[width=0.5\linewidth]{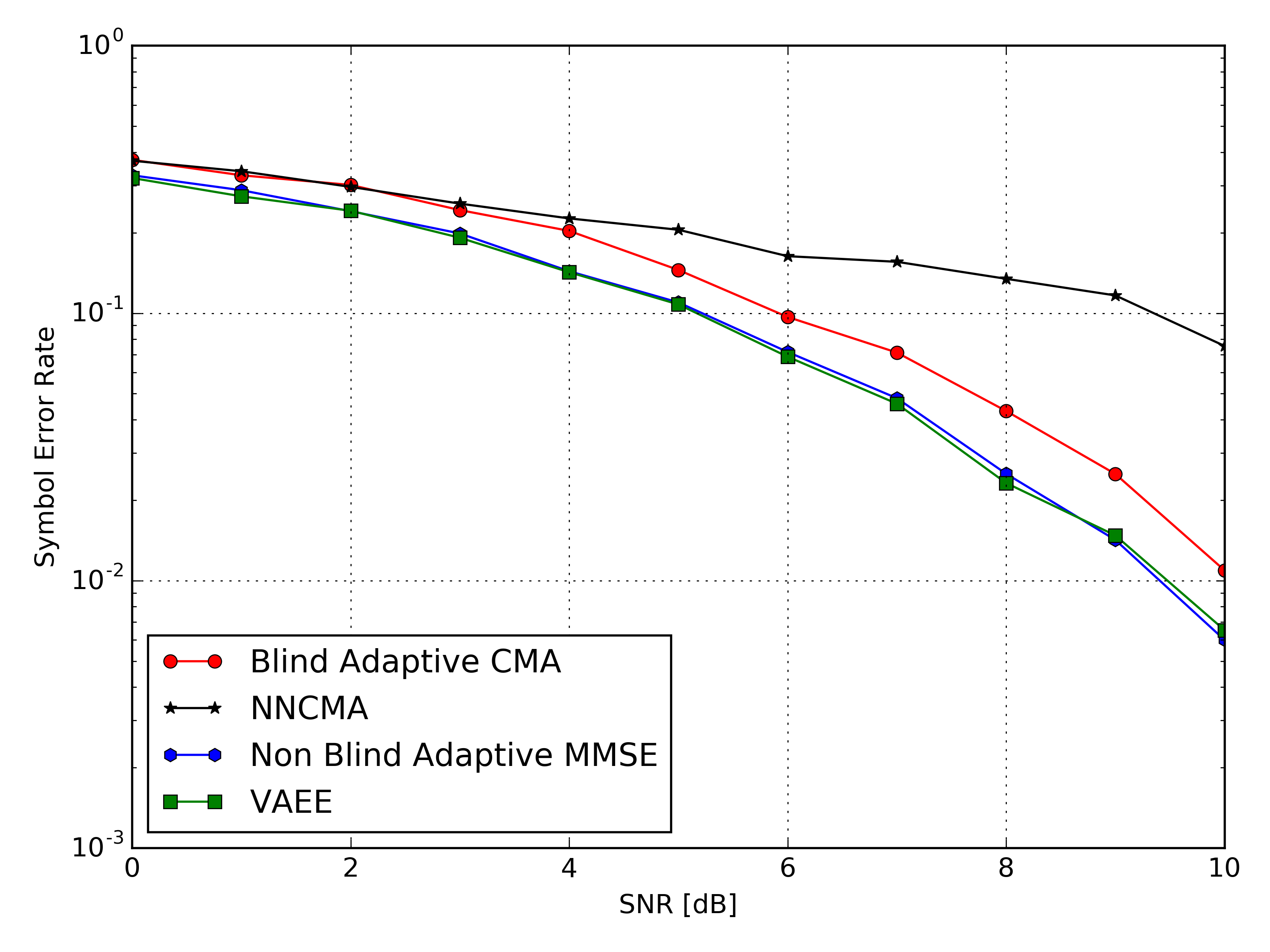}
	\caption{SER vs. SNR for the equalization algorithms. The channel is $\bh_{1}$.}
	\label{fig:snr1}
\end{figure}

\begin{figure}[]
	\centering
	\includegraphics[width=0.5\linewidth]{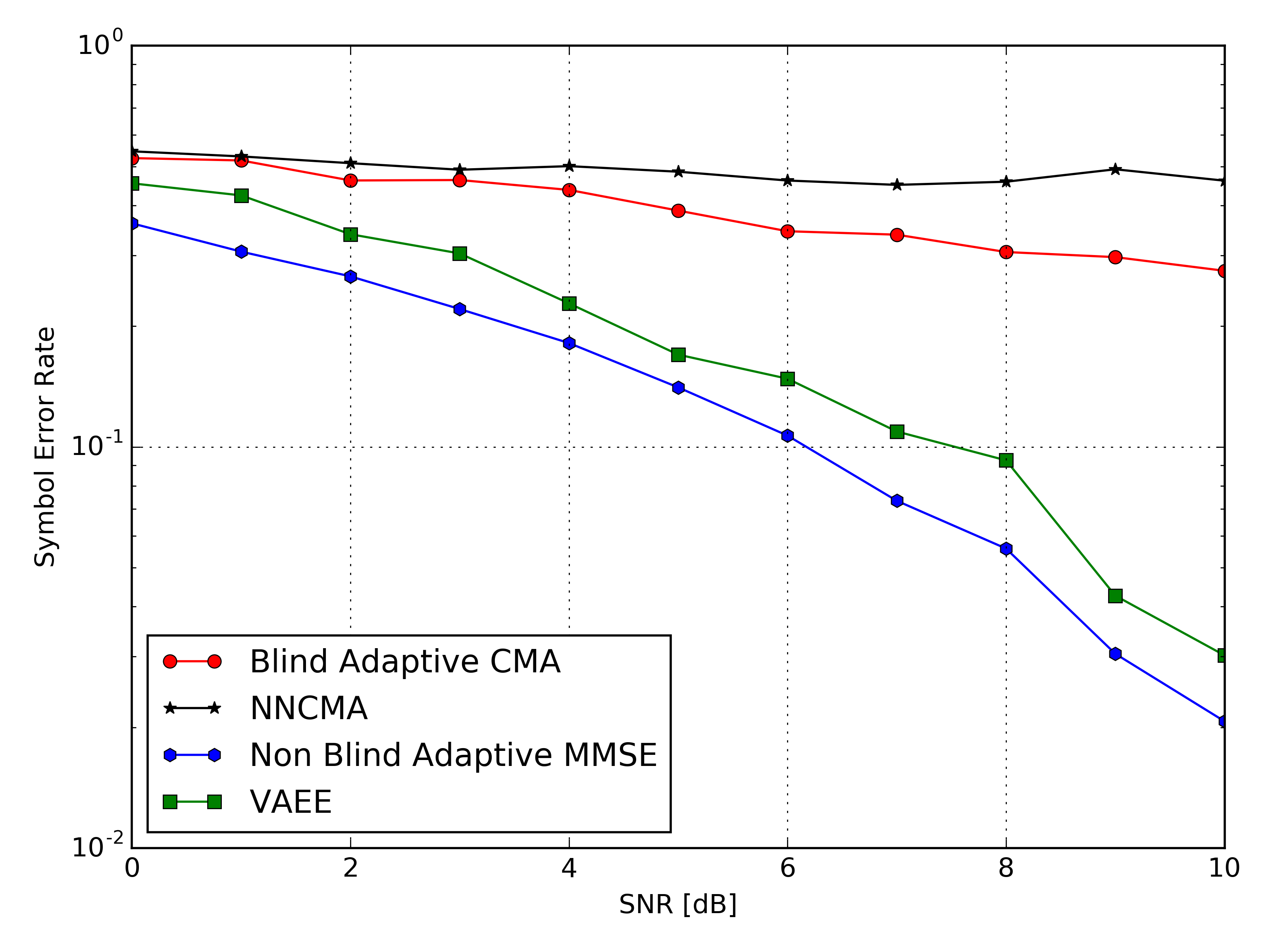}
	\caption{SER vs. SNR for the equalization algorithms. The channel is $\bh_{2}$.}
	\label{fig:snr2}
\end{figure}

\begin{figure}
	\centering
	\includegraphics[width=0.5\linewidth]{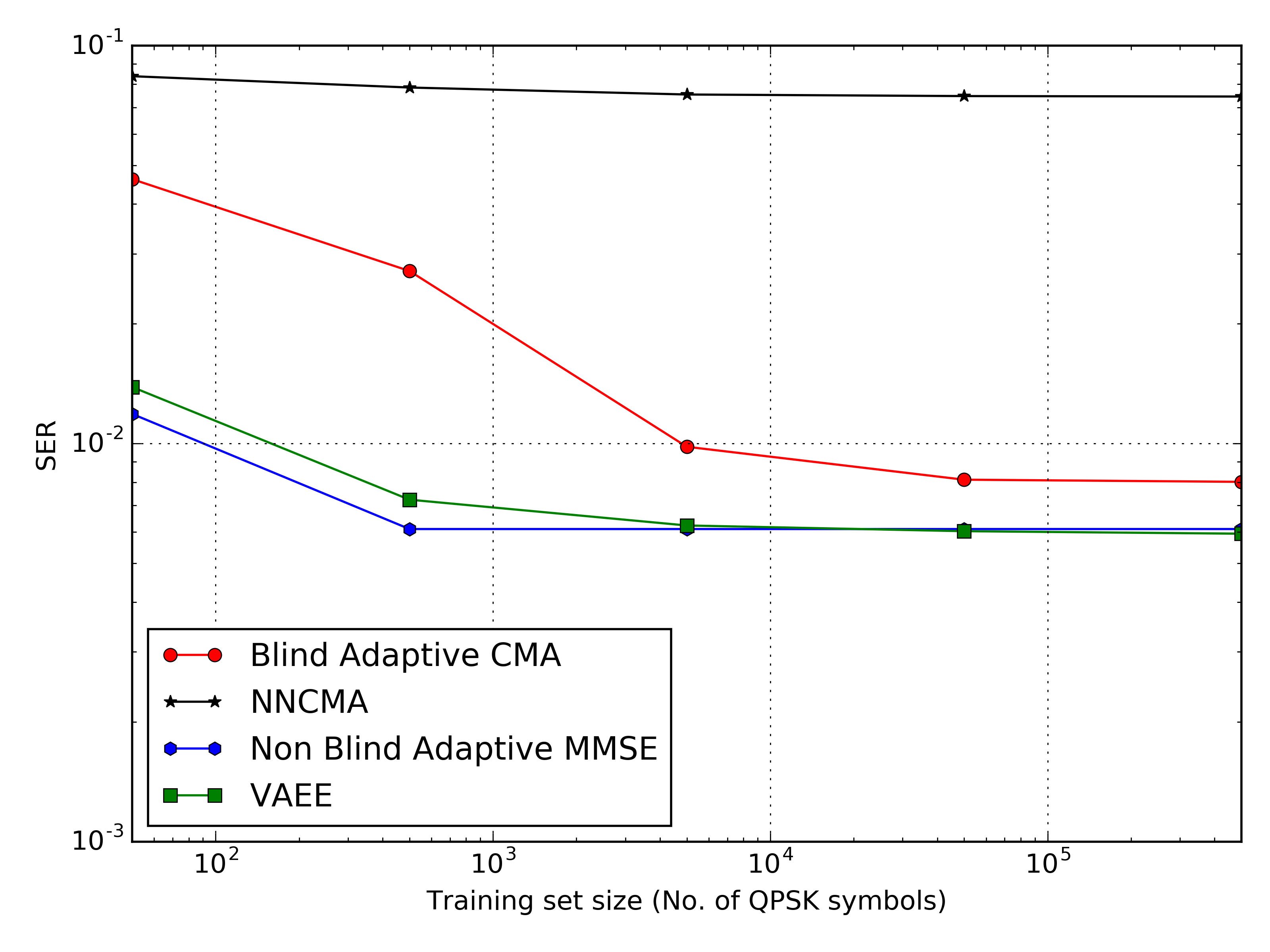}
	\caption{SER vs. number of training symbols for the equalization algorithms. $\text{SNR}=10$dB. The channel is $\bh_1$.}
	\label{fig:data10}
\end{figure}

We start by reporting results for noisy linear ISI channels under QPSK modulation, without using coding information.
Our algorithm was compared with the adaptive CMA \cite{abrar2010adaptive}, and with the neural network CMA (NNCMA) \cite{you1998nonlinear} blind equalization algorithms.
In addition, we compared the performance to the adaptive linear MMSE \cite{gong2010adaptive} non-blind equalizer that observes the actual transmitted sequence.
The baseline algorithms are on-line algorithms. This means that for each incoming data sample, a single update is made and this sample is not used any longer. In order to improve the performance of the baseline algorithms, we have modified them so that after using the entire given block of data, based on which we learn the channel, we start a new training round on the given block of data, and repeat this process until convergence.
In our first experiments in this subsection, reported in Figs. \ref{fig:snr1}, \ref{fig:snr2} and \ref{fig:data10}, we used the Adam optimization algorithm \cite{kingma2014adam} to minimize our loss function.
For all experiments in the uncoded case, and all blind equalization methods, we note that one can recover the transmitted bits only up to some unknown delay and rotation of the constellation, which for QPSK means that we need to examine four different possible rotations ($0\degree,90\degree,180\degree,270\degree)$.
For each SNR point in the experiments in this subsection, we obtained the symbol error rate (SER) by averaging over the SER values corresponding to 20 independent sets of training and test data sequences. Each training sequence contained the channel observations corresponding to $L$ (unknown) random QPSK symbols (with $L$ varying between experiments as described below), and each test data sequence contained the channel observations corresponding to $K=\text{10,000}$ random QPSK data symbols.
As an alternative we could have tested the results on the same data used for training (for this data too, we do not know the actual transmitted symbol at the receiver since we assume an unsupervised setup). When doing so, the SER remained essentially the same as for the independent test data in all the experiments.
In calculating the SER we took into account all possible rotations and delays.

In all our experiments, we used the same convolutional neural network VAE-encoder architecture in Fig. \ref{fig:architecture}, with a filter with five complex coefficients in the first layer, and a filter with two complex coefficients in the second layer. Hence, the total number of free parameters in the model was only $M+15$ ($M$ channel impulse response parameters in the VAE-decoder, 1 parameter representing the noise variance, and 14 ($2\times(5+2)$) real parameters in the convolutional neural network VAE-encoder).

In our first set of experiments,
we compared our model to the baseline algorithms at various noise levels, using the following non-minimum phase channels taken from \cite{ranhotra2017performance,fang1999blind} 
\begin{equation*}
\begin{split}
\bh_1 =& [0.0545+0.05j, 0.2832-0.11971j, -0.7676+0.2788j,\\& -0.0641-0.0576j, 0.0466-0.02275j]\\
\bh_2 =& [0.0554+0.0165j, -1.3449-0.4523j,\\ & 1.0067+1.1524j, 0.3476+0.3153j]
\end{split}
\end{equation*}
We generated $L=2000$ random QPSK symbols as the training sequence.
Then we applied convolution with the channel impulse response, and added white Gaussian noise at a signal to noise ratio (SNR) in the range $0$dB -- $10$dB. The SNR is defined by
$
\text{SNR} \defined 20\log_{10} \left( ||\bx*\bh|| / ||\bw|| \right)
$.
To train the model, for each update step, we sampled from the training set a mini-batch of a single sub-sequence of length $N=128$.
Figs. \ref{fig:snr1} and \ref{fig:snr2} present SER results for $\bh_1$ and $\bh_2$ respectively.
As can be seen, the new VAEE significantly outperforms the baseline blind equalizers, and is quite close to the performance of the non-blind adaptive linear MMSE equalizer.

In our following experiment, we compared the SER of the equalization algorithms as the number of training symbols varied from $L=\text{50}$ to $L=\text{500,000}$. For each update step we sampled from the training set a mini-batch of a single sub-sequence of length $N=\min\left(128,L\right)$. We used the channel impulse response $\bh_1$ above.
Fig. \ref{fig:data10} presents the results for SNR=10dB.
Again, the new VAEE algorithm significantly outperforms the baseline blind equalization algorithms. The results show that the VAEE enables faster channel acquisition compared to the other blind equalization algorithms. 


Further experiments for uncoded data are reported in \cite{8403666}.

\begin{figure}[]
\centering
\includegraphics[width=0.5\linewidth]{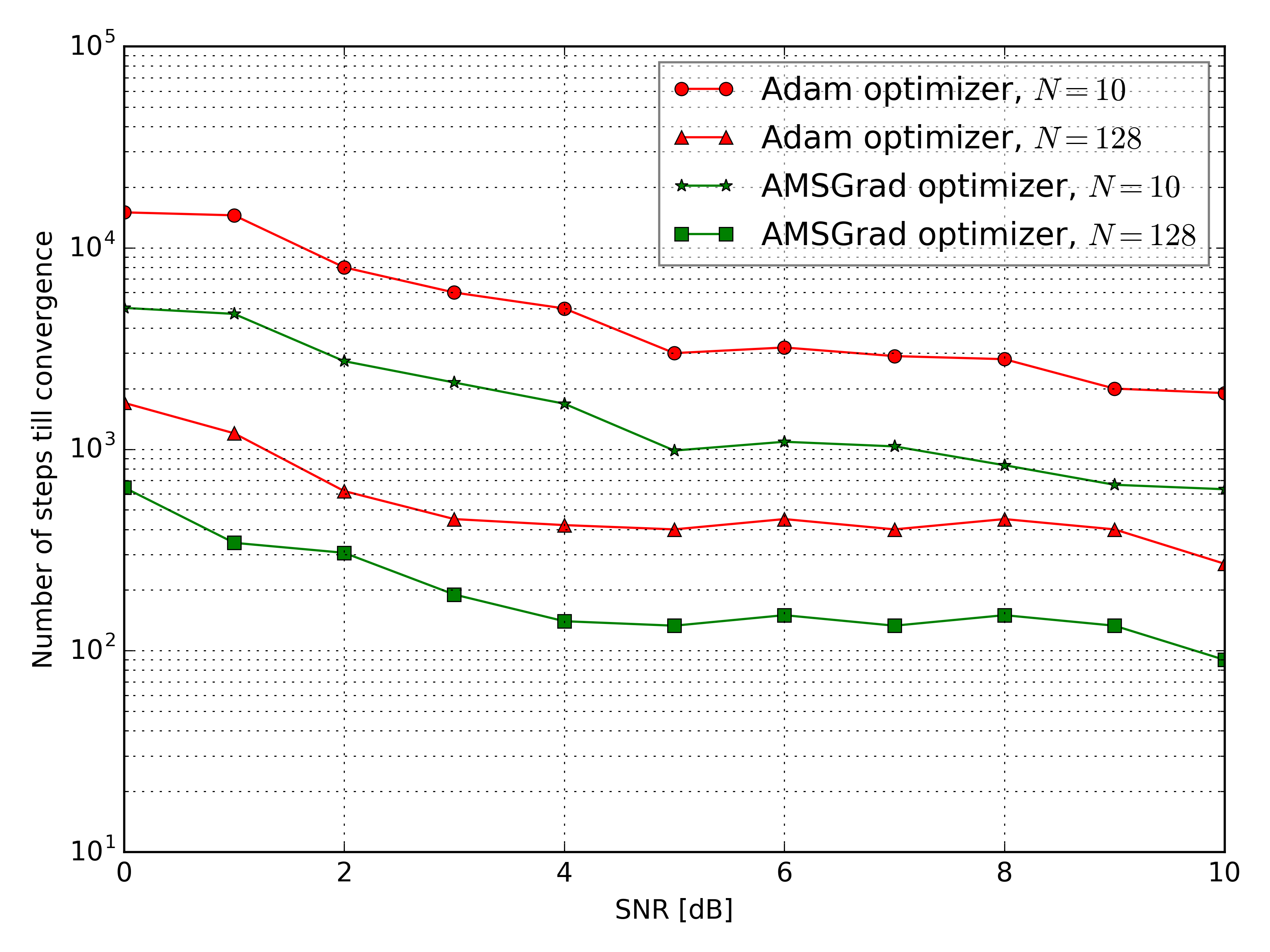}
\caption{Number of parameter updates vs. SNR for different $N$. The channel is $\bh_1$.}
\label{fig:snr_iters}
\end{figure}

For maintaining good performance but faster convergence time, we re-ran simulations from \cite{8403666} using a variety of gradient descent-based optimizers.
In Fig. \ref{fig:snr_iters} we report on the number of parameter updates required for convergence of the VAEE algorithm when using the channel $\bh_1$. To train the model, we sampled a mini-batch of a single sub-sequence of length $N\in\{10,128\}$ out of the given training symbols.
Then we let the algorithm train until convergence was achieved.
As presented in Fig. \ref{fig:snr_iters}, the AMSGrad optimizer \cite{reddi2018convergence} leads to a significant speedup in the training of our model compared to the Adam algorithm \cite{kingma2014adam}.
The bit error rate (BER) and SER performance did not change much between optimizers.
Thus, we used the AMSGrad optimizer in our subsequent simulations.

\subsection{Linear channels, coded data} \label{sec:sim_lin_cod}
In all our experiments with coded data (this subsection and the following one) we assume the availability of only the channel observations corresponding to a single transmitted LDPC codeword (i.e., we do not have additional training data). As explained above, our goal is to reconstruct the transmitted message from this data alone without knowing the channel parameters.
As an upper bound on the performance, we compared our results to the non-blind turbo equalization algorithm \cite{douillard1995iterative,koetter2004turbo} which knows the true channel. This algorithm applies the BCJR algorithm \cite{bahl1974optimal} and the BP algorithm iteratively.

We then implemented and compared our algorithm to the (blind) EM algorithm \cite{dempster1977maximum} for a noisy linear ISI channel, taking into account the coding information similarly to \cite{1510991}. In every iteration, this \emph{turbo EM} algorithm re-estimates the ISI channel impulse response and the noise variance. It applies the generalized BCJR algorithm \cite{gunther2007generalized} and the BP algorithm iteratively as in the turbo equalization algorithm.
As explained in \cite{gunther2007generalized}, the accurate implementation of the EM requires the computation of the posterior joint expectation $\E\left[x_i,x_j\right|\by]$ of two transmitted symbols. In the approximate EM algorithm for a noisy ISI channel proposed in \cite{kaleh1994joint}, the approximation $\E\left[x_i x_j\right|\by] \approx \E\left[x_i\right|\by]\cdot \E\left[x_j\right|\by]$ is used. Under this approximation, the standard BCJR algorithm is sufficient to implement EM estimation. To improve results, a generalized BCJR is derived in \cite{gunther2007generalized} to compute $\E\left[x_i, x_j\right|\by]$ accurately.
We have also implemented and evaluated channel estimation with the least squares method as described in \cite{ghosh1992maximum}. These results are not shown since in all our experiments the EM algorithm outperformed the least squares estimation method.
As noted above, the turbo equalization algorithm is channel informed, i.e., it knows the true ISI channel impulse response and noise variance. On the other hand, our turbo VAEE algorithm and the turbo EM algorithm are blind. They perform unsupervised joint estimation of the channel coefficients, noise variance and the transmitted codeword.

In all our experiments with coded data, we used the same convolutional neural network VAE-encoder architecture in Fig. \ref{fig:architecture_VAD}. We used a filter with $10$ coefficients in the first layer, and a filter with $5$ coefficients in the second layer. Hence, the total number of free parameters in the model was $M+16$ ($M$ parameters for the channel impulse response, $1$ parameter for the channel noise variance, and $15$ parameters for the convolutional neural network VAE-encoder).
In the turbo VAEE experiments we first applied standalone VAEE (with a Gallager loss term) for $I$ iterations, and only then started turbo mode where we apply $T$ external iterations, each consisting of one VAEE iteration and $B$ BP iterations. The values of the hyper-parameters used are summarized in Table \ref{table:parameters}.
\begin{table}[htbp]
\caption{Values of the hyper-parameters. \vspace{-0.3cm}} 
\begin{center}
\scalebox{1.0}
{
\begin{tabular}{|| c | c | c ||}
  \hline 
  Variable & Definition & Value \\ \hline \hline 
      $T$ &  Total number of training iterations & $80$\\ \hline 
    $\text{lr}$ &  Learning rate for VAEE training & $10^{-1}$\\ \hline 
    $\lambda$ &  Weight in \eqref{eq:lossVAD}  & $0.7$\\ \hline 
    $\eta$ &  Attenuation factor in Section V.B & $0.1$\\ \hline 
    $\alpha$ &  Coefficient in \eqref{eq:pjx_1} & $0.2$\\ \hline 
       $B$ & Number of BP iterations   & $15$ \\  
         &  after each VAEE iteration  &  \\ \hline
           $I$ & Number of VAEE iterations & $50$ \\  
         &  before applying turbo mode &  \\ \hline
\end{tabular}
}
\end{center}\vspace{-0.3cm}
\label{table:parameters}
\end{table}

In the results presented in the following experiments, for each SNR point, we display the average BER obtained by repeating the following basic experiment: At the transmitter we encode one codeword (unknown at the receiver) and transmit it over the channel. At the receiver we apply the decoding algorithm on the channel observations to decode the unknown codeword. Then we measure the decoding BER, defined as the fraction of erroneous decoded bits out of the $N$ code bits. Now, in order to obtain a statistically reliable BER estimate, we repeat this basic experiment several times, each time with a different randomly chosen codeword and a different randomly chosen channel noise realization, and calculate the average BER reported in the figures.
We evaluated the various decoding algorithms at various noise levels, using two LDPC codes from the Wimax IEEE 802.16e standard. The parity check matrices are taken from \cite{channelcodes}. The first code has blocklength $N_{1}=576$ and the second has blocklength $N_{2}=2304$. The rate of both codes is $3/4$. The following causal non-minimum phase channel impulse responses were used to simulate the ISI,
\begin{equation*}
\begin{split}
\tilde{\bh}_{1} =& [0.2, 0.9, 0.3]\\
\tilde{\bh}_{2} =& [0.2, 0.9, 0.3, 1.0]\\
\tilde{\bh}_{3} =& [0.16, 0.545, -0.672, 0.256, 0.095, -0.389]
\end{split}
\end{equation*}
Channel $\tilde{\bh}_{3}$ is taken from \cite{salamanca2010channel}.



\begin{figure}[]
\centering
\begin{subfigure}[h]{0.49\linewidth}
\includegraphics[width=1\linewidth]{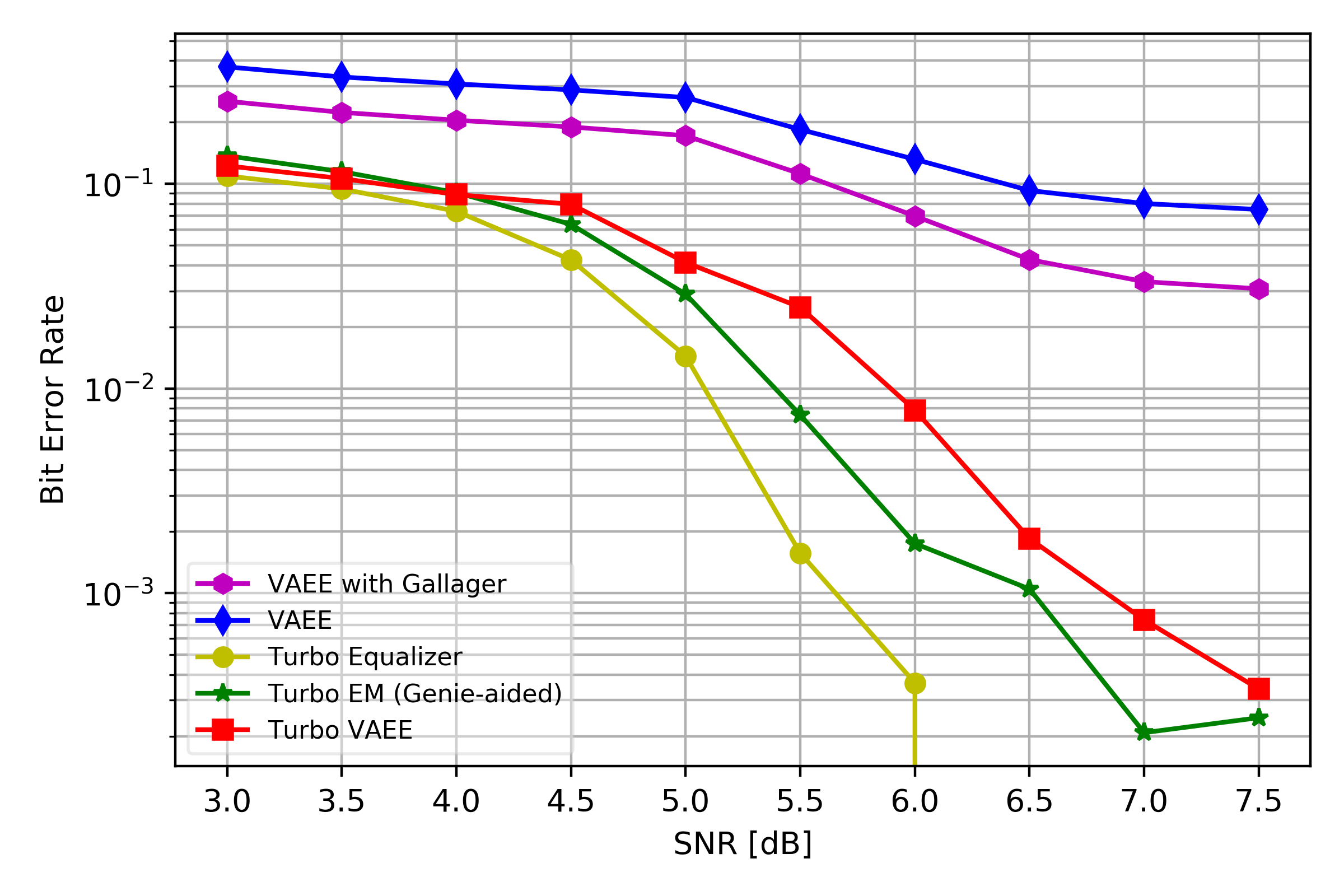}
\caption{LDPC code with blocklength $N_{1}$}
\end{subfigure}
\hfill
\begin{subfigure}[h]{0.49\linewidth}
\includegraphics[width=1\linewidth]{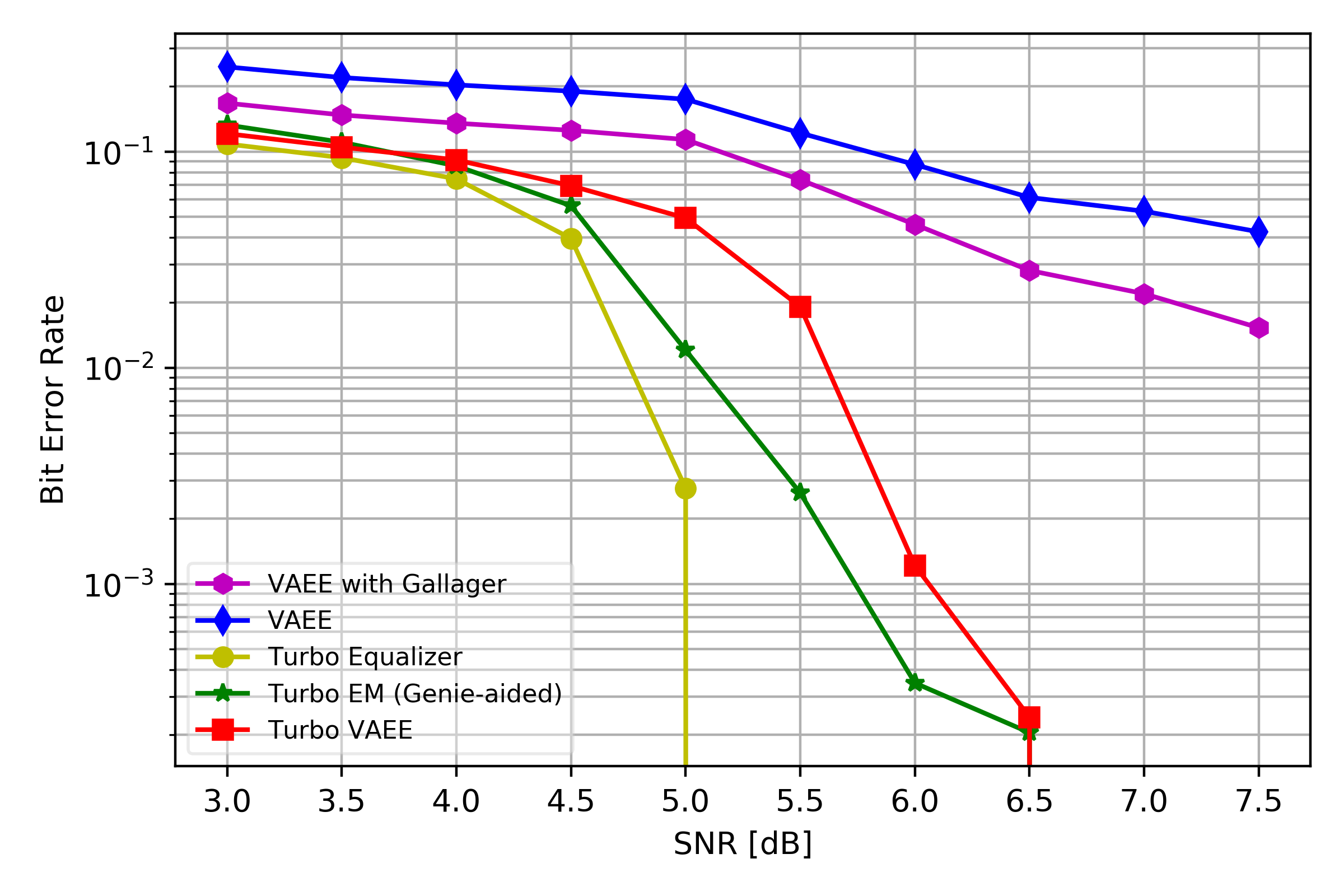}
\caption{LDPC code with blocklength $N_{2}$}
\end{subfigure}
\caption{BER vs. SNR for the blind decoding algorithms and for the channel informed turbo equalizer. 
The channel is $\tilde{\bh}_{1}$.}
\label{fig:snr_h1}
\end{figure}

\begin{figure}[]
\centering
\begin{subfigure}[h]{0.49\linewidth}
\includegraphics[width=1\linewidth]{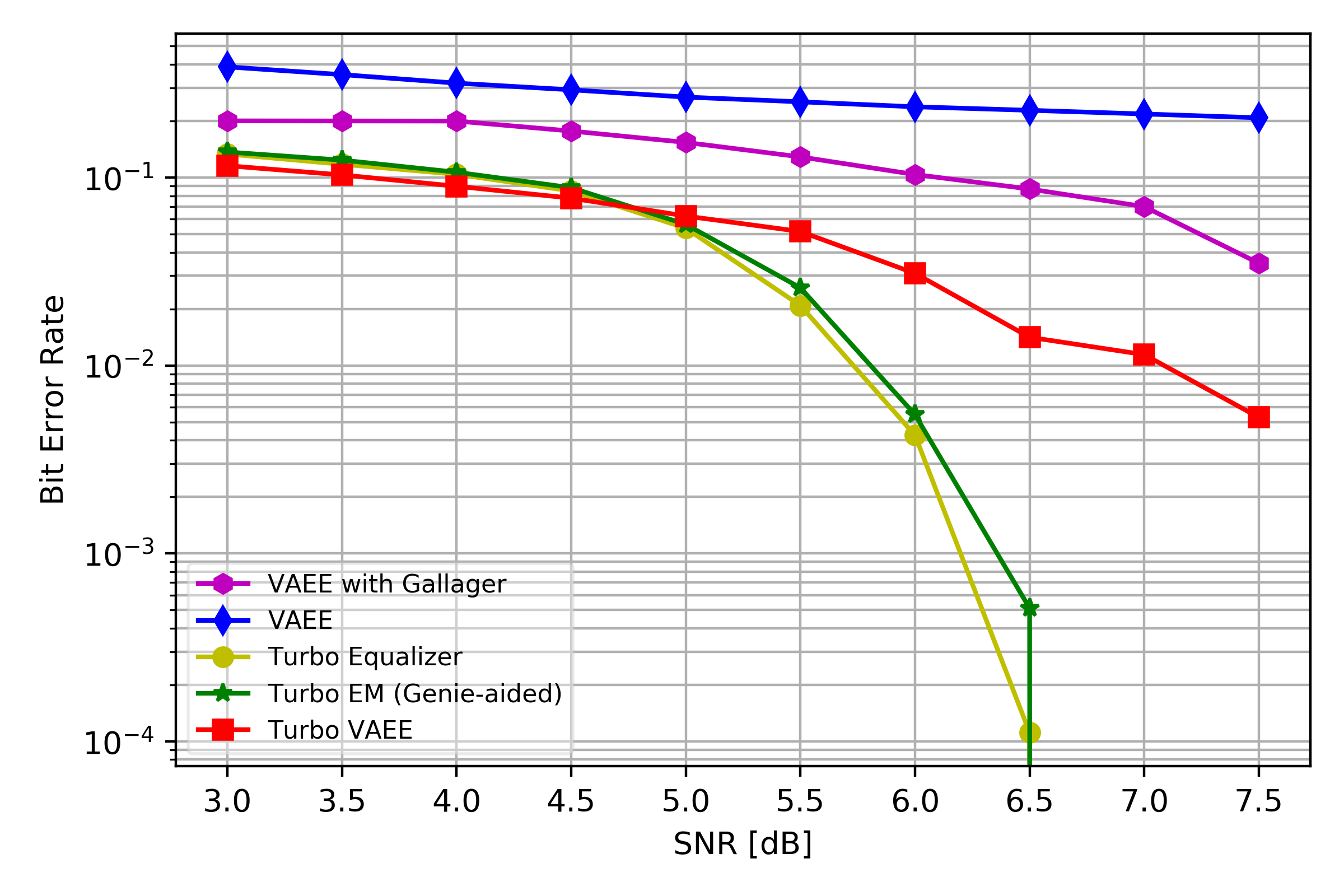}
\caption{LDPC code with blocklength $N_{1}$}
\end{subfigure}
\hfill
\begin{subfigure}[h]{0.49\linewidth}
\includegraphics[width=1\linewidth]{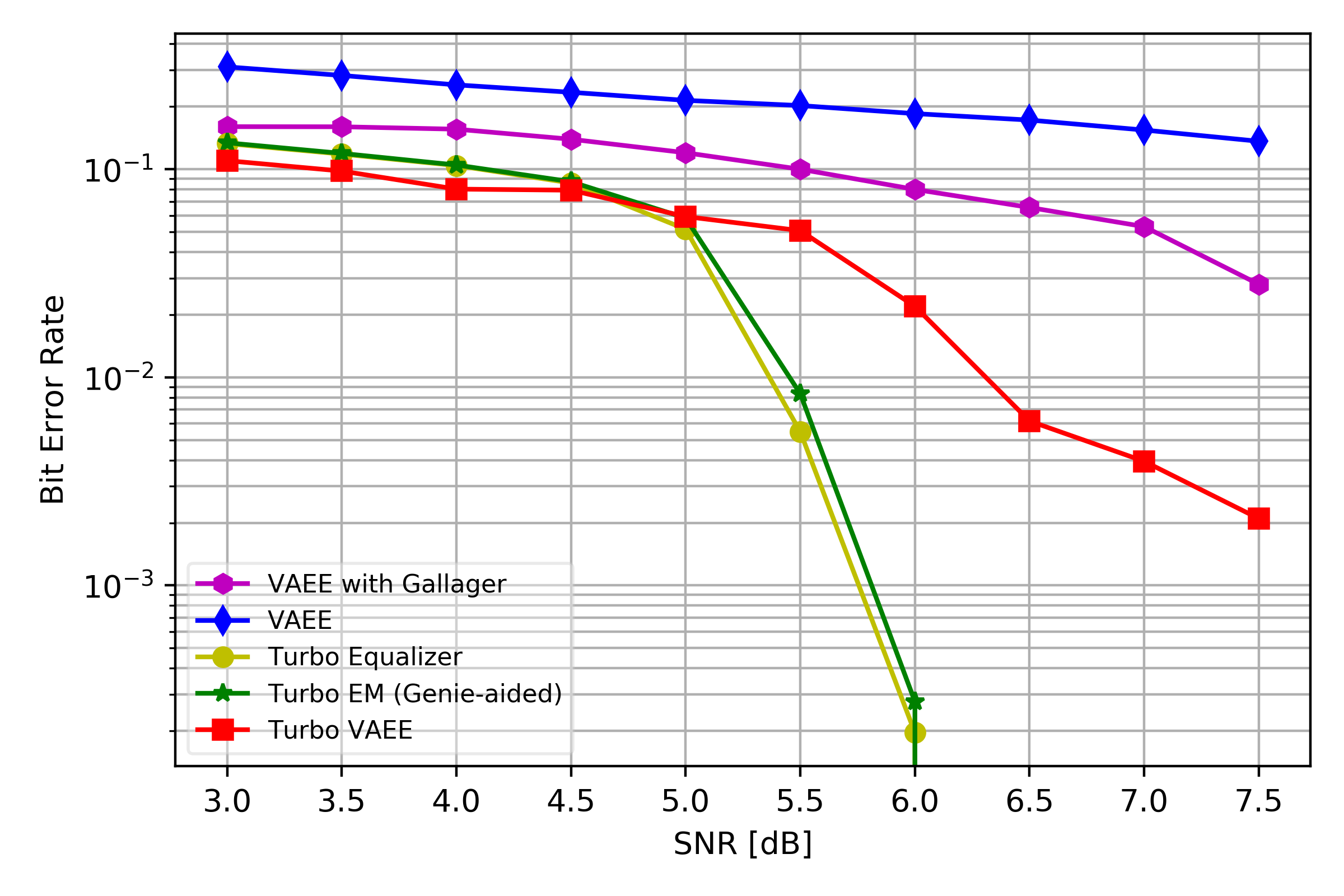}
\caption{LDPC code with blocklength $N_{2}$}
\end{subfigure}
\caption{BER vs. SNR for the blind decoding algorithms and for the channel informed turbo equalizer. The channel is $\tilde{\bh}_2$.}
\label{fig:snr_h2}
\end{figure}

\begin{figure}[]
\centering
\begin{subfigure}[h]{0.49\linewidth}
\includegraphics[width=1\linewidth]{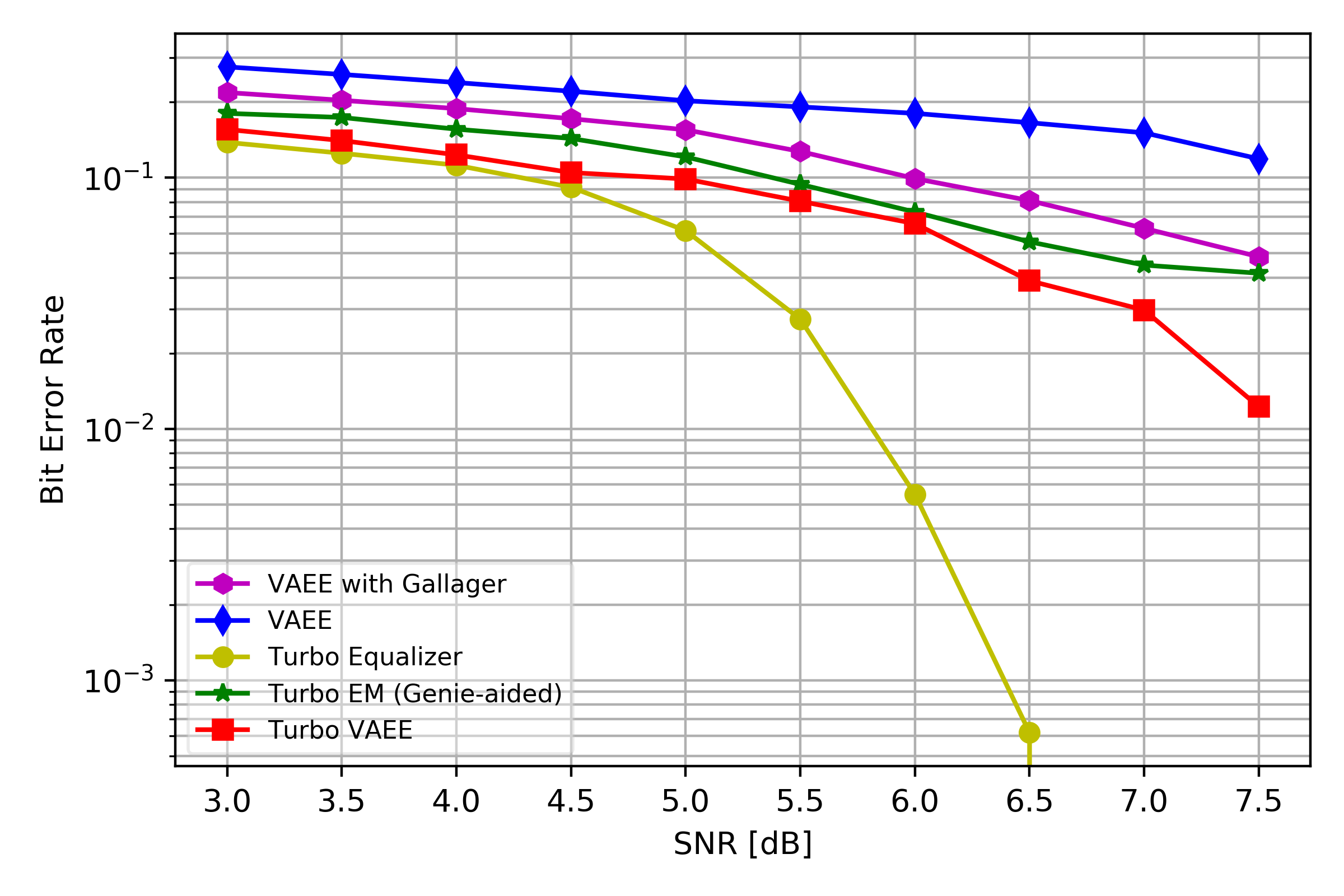}
\caption{LDPC code with blocklength $N_{1}$}
\end{subfigure}
\hfill
\begin{subfigure}[h]{0.49\linewidth}
\includegraphics[width=1\linewidth]{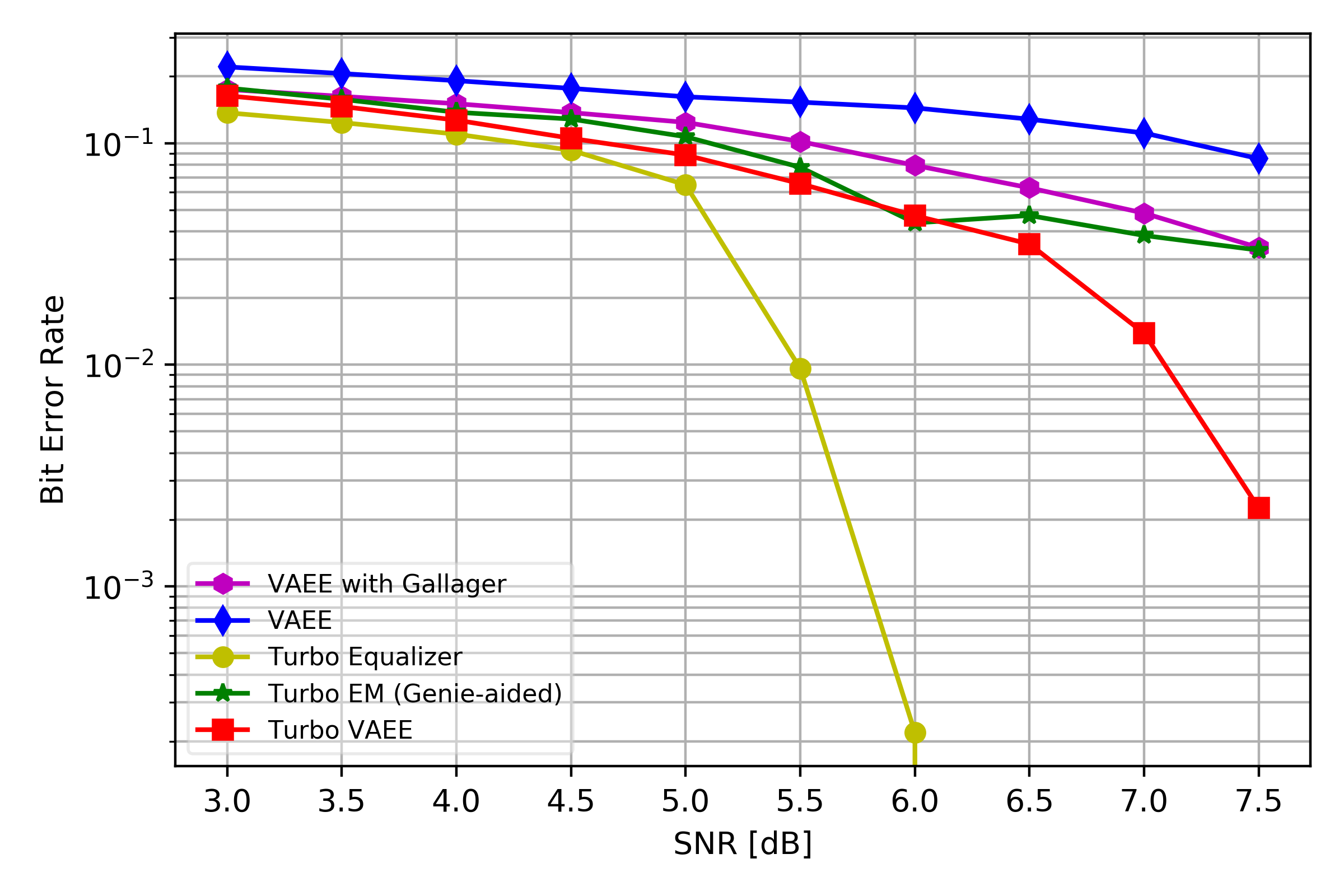}
\caption{LDPC code with blocklength $N_{2}$}
\end{subfigure}
\caption{BER vs. SNR for the blind decoding algorithms and for the channel informed turbo equalizer. The channel is $\tilde{\bh}_{3}$.}
\label{fig:snr_h3}
\end{figure}

For reference, we have computed the Shannon threshold SNR for which channel capacity is equal to the code rate used ($3/4$).
The obtained thresholds for $\tilde{\bh}_{1}$, $\tilde{\bh}_{2}$ and $\tilde{\bh}_{3}$ were $2.84\text{dB}$, $2.95\text{dB}$ and $3.0\text{dB}$ respectively.
The capacity of the noisy ISI channel was computed using the water-filling algorithm \cite{cover_book}. We have not imposed BPSK modulation in the computation of the capacity. Hence, the actual capacity under BPSK is smaller and the corresponding Shannon threshold SNR is higher than reported.
Figs. \ref{fig:snr_h1}, \ref{fig:snr_h2} and \ref{fig:snr_h3} present BER results for the channels $\tilde{\bh}_1$, $\tilde{\bh}_2$ and $\tilde{\bh}_3$, respectively, for the two codes with blocklengths $N_{1}$ and $N_{2}$.
The blind decoding algorithms that were examined include standalone VAEE, standalone VAEE with a Gallager loss term as described in Section \ref{sec:VAE_Gal_lemma}, turbo VAEE with a Gallager loss term $\cL_G$ component and BP iterations as described in Sections \ref{sec:VAE_Gal_lemma} and \ref{Using_Turbo_mode}, and turbo EM. As a practical upper bound on the achievable performance we also plot the BER of the channel informed non-blind turbo equalizer.
Naive implementation of turbo EM did not work well. To boost the performance we had to first apply EM without incorporating code information (i.e., without using the BP algorithm). Then, in a second stage, the BP could be incorporated to the scheme. However, since the first EM stage does not use code information, it has an inherent ambiguity regarding the polarity of the channel impulse response (if $\hat{\bh}$ is the estimated response, then $-\hat{\bh}$ is also a good possibility). Hence, we need to carry out the second EM stage using both $\hat{\bh}$ and $-\hat{\bh}$ and choose that option with minimum BER. However, in reality we cannot compute the error rate. Furthermore, even though we assumed that the length of the channel impulse response, $M$, is known, so that the optimal estimated response is close to $\bh$ or $-\bh$, the EM algorithm occasionally converged to a local minimum which was close to a shift left or a shift right of $\bh$ (or $-\bh$). Hence to obtain good results, in all our simulations with EM we tested all 6 possibilities after the first EM stage (the two possible polarities and the three possibilities for shift left by 1, shift right by 1 and no shift). For each possibility we applied the second EM stage with BP iterations incorporated, and finally we picked the option that yielded minimum BER. This is a genie-aided EM that cannot be used in practice. In order to make it a practical algorithm, one could incorporate CRC bits within the transmitted message bits, but this would incur rate loss\footnote{We tried using the Hamming weight of the syndrome of the decoded codeword instead of using BER, but this did not work well.}
On the other hand, the turbo VAEE algorithm did not have this problem, due to the Gallager loss term, and we used a single run (rather than 6 runs) for each simulation. As can be seen, the new blind turbo VAEE algorithm is worse than the genie-aided turbo EM for $\tilde{\bh}_1$ and $\tilde{\bh}_2$, but is better even than genie-aided turbo EM for $\tilde{\bh}_3$, which does not work well for this channel with the longer impulse response. A practical (non-genie-aided) EM algorithm was significantly inferior to turbo VAEE for all channels.
Standalone VAEE and VAEE with Gallager loss performed much worse then the full turbo VAEE algorithm (that incorporates the Gallager loss).
However, these algorithms require much less computations compared to turbo VAEE.

\subsection{Nonlinear channels} \label{sec:sim_nonlin}
We simulated the nonlinear channels as proposed in \cite{patra2009nonlinear, ye2017initial, 8491056, olmos2010joint, mitchinson2002digital},
\begin{equation*}
\begin{split}
g_{1}\left(a_n\right) =& \tanh{\left(a_n\right)}\\
g_{2}\left(a_n\right) =& a_{n}+0.2a_{n}^{2}-0.1a_{n}^{3}\\
g_{3}\left(a_n\right) =& a_{n}+0.2a_{n}^{2}-0.1a_{n}^{3} +0.5\cos{\left(\pi a_n\right)}
\end{split}
\end{equation*}
where we define $a_{n}\defined\sum_{k} x_k h_{n-k}$.
It should be noted that $g_2(\cdot)$ and $g_3(\cdot)$ represent an amplifier working in saturation.

As a practical upper bound on the performance, we compared our results to a non-blind turbo equalization algorithm which knows the true channel impulse response $\bh$, the noise variance $\sigma_w^2$ and the nonlinearity $g(\cdot)$. This algorithm uses a modified BCJR algorithm which is very similar to the standard BCJR algorithm \cite{bahl1974optimal}, except that after we have computed the convolution $a_n$ we apply the nonlinear function $g(\cdot)$. That is, we compute $g(a_n)$ and proceed as before.
We also compared our turbo VAEE algorithm to the turbo EM algorithm that was used in the previous subsection. This algorithm ignores the nonlinearity. To the best of our knowledge there does not exist any other baseline blind estimation algorithm for the nonlinear case that we could compare with. 

\begin{figure}[]
	\centering
	\begin{subfigure}[h]{0.49\linewidth}
		\includegraphics[width=1\linewidth]{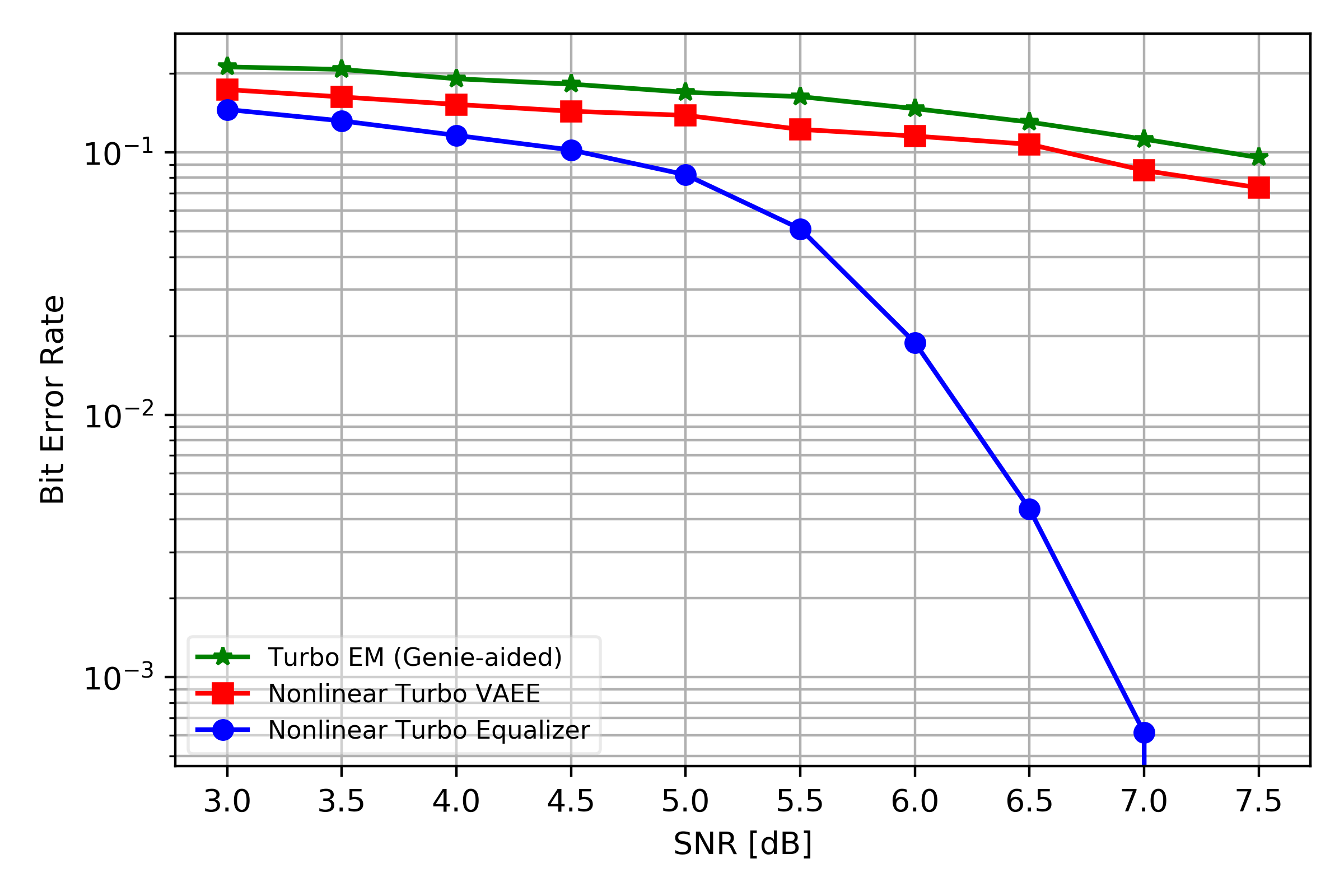}
		\caption{LDPC code with blocklength $N_{1}$}
	\end{subfigure}
	\hfill
	\begin{subfigure}[h]{0.49\linewidth}
		\includegraphics[width=1\linewidth]{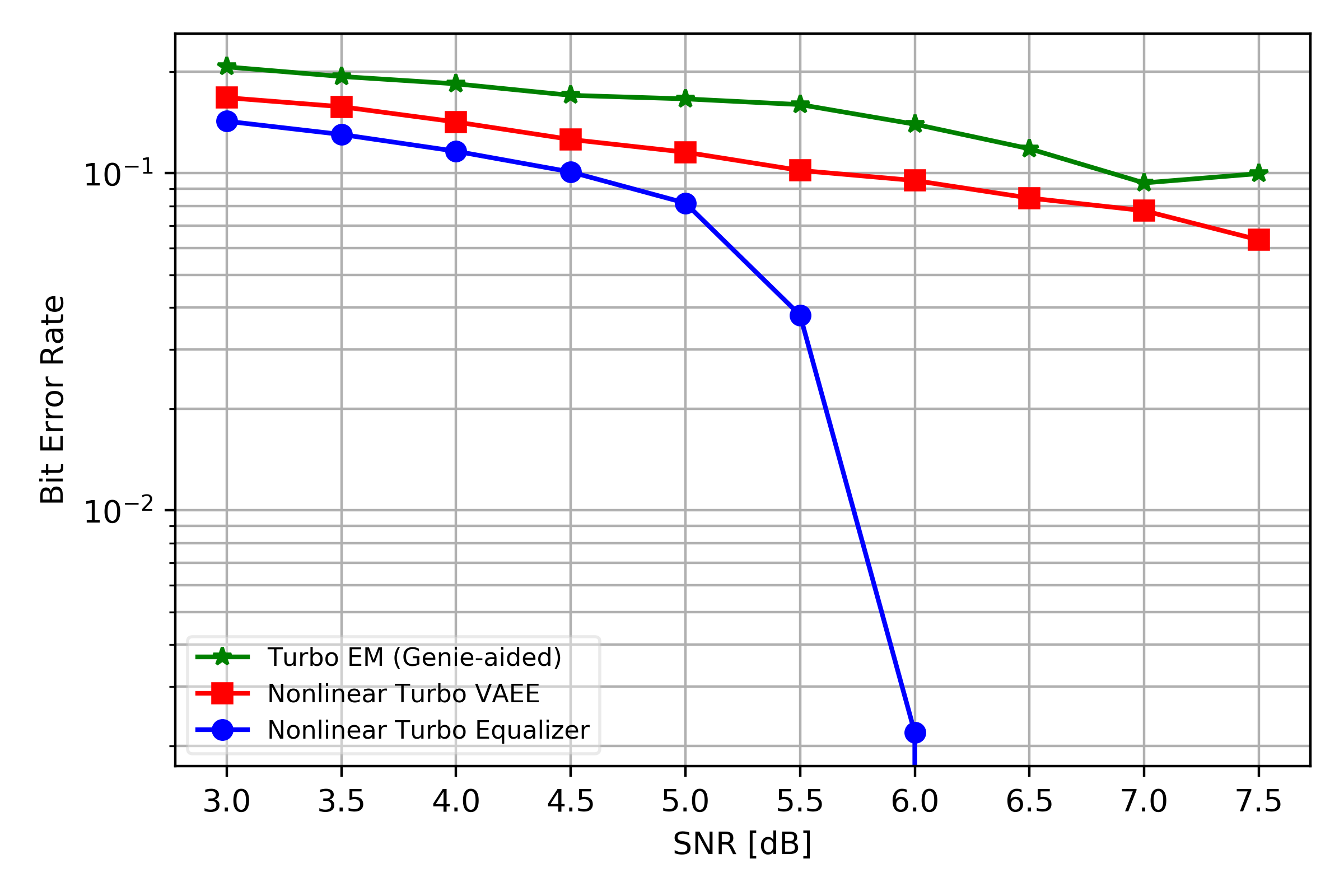}
		\caption{LDPC code with blocklength $N_{2}$}
	\end{subfigure}
	\caption{BER vs. SNR for the blind decoding algorithms and for the channel informed turbo equalizer. 
		The channel is $\tilde{\bh}_3$. The nonlinearity is $g_{1}(\cdot)$.}
	\label{fig:snr_g1}
\end{figure}

\begin{figure}[]
	\centering
	\begin{subfigure}[h]{0.49\linewidth}
		\includegraphics[width=1\linewidth]{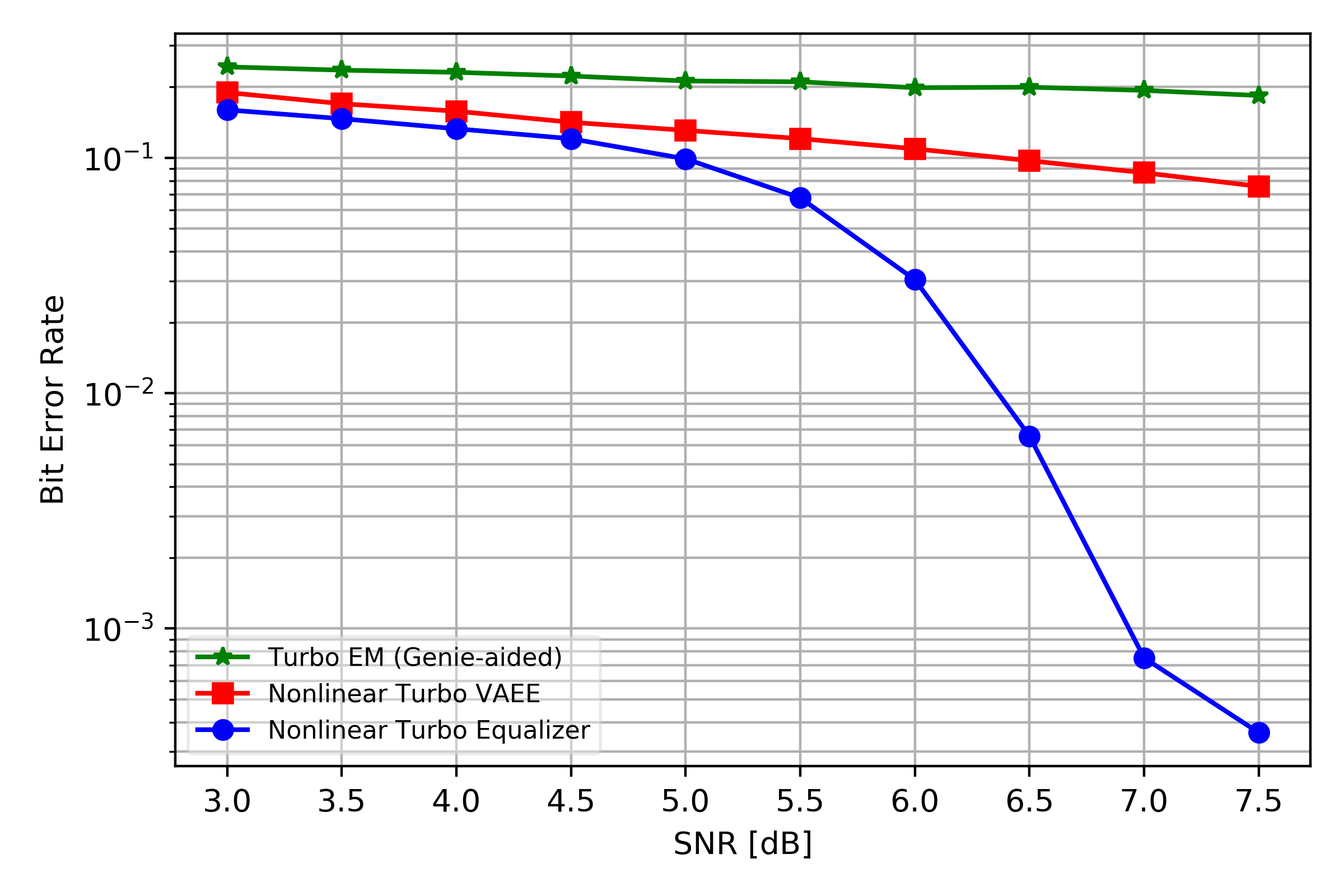}
		\caption{LDPC code with blocklength $N_{1}$}
	\end{subfigure}
	\hfill
	\begin{subfigure}[h]{0.49\linewidth}
		\includegraphics[width=1\linewidth]{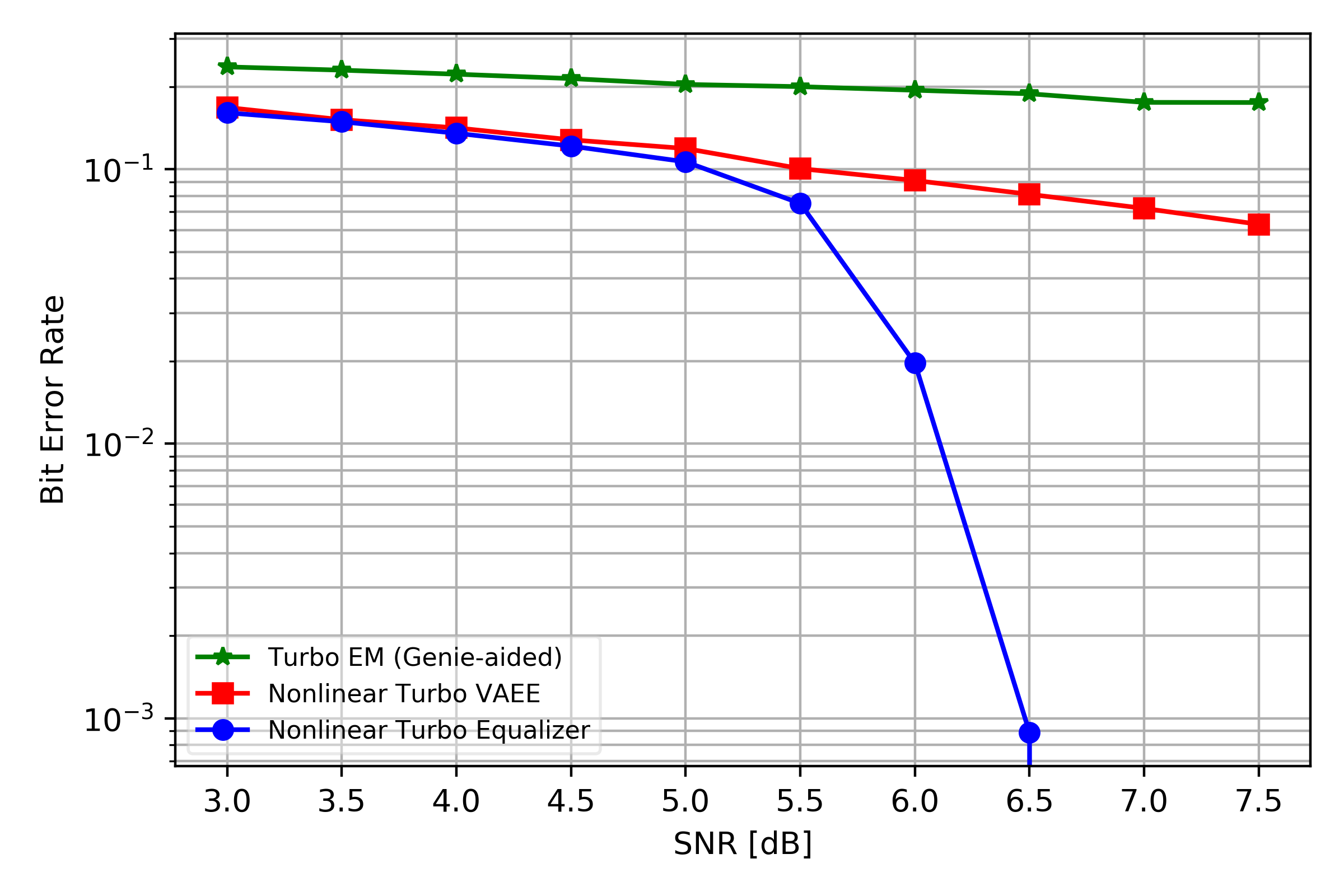}
		\caption{LDPC code with blocklength $N_{2}$}
	\end{subfigure}
	\caption{BER vs. SNR for the blind decoding algorithms and for the channel informed turbo equalizer. The channel is $\tilde{\bh}_3$. The nonlinearity is $g_{2}(\cdot)$.}
	\label{fig:snr_g2}
\end{figure}

\begin{figure}[]
	\centering
	\begin{subfigure}[h]{0.49\linewidth}
		\includegraphics[width=1\linewidth]{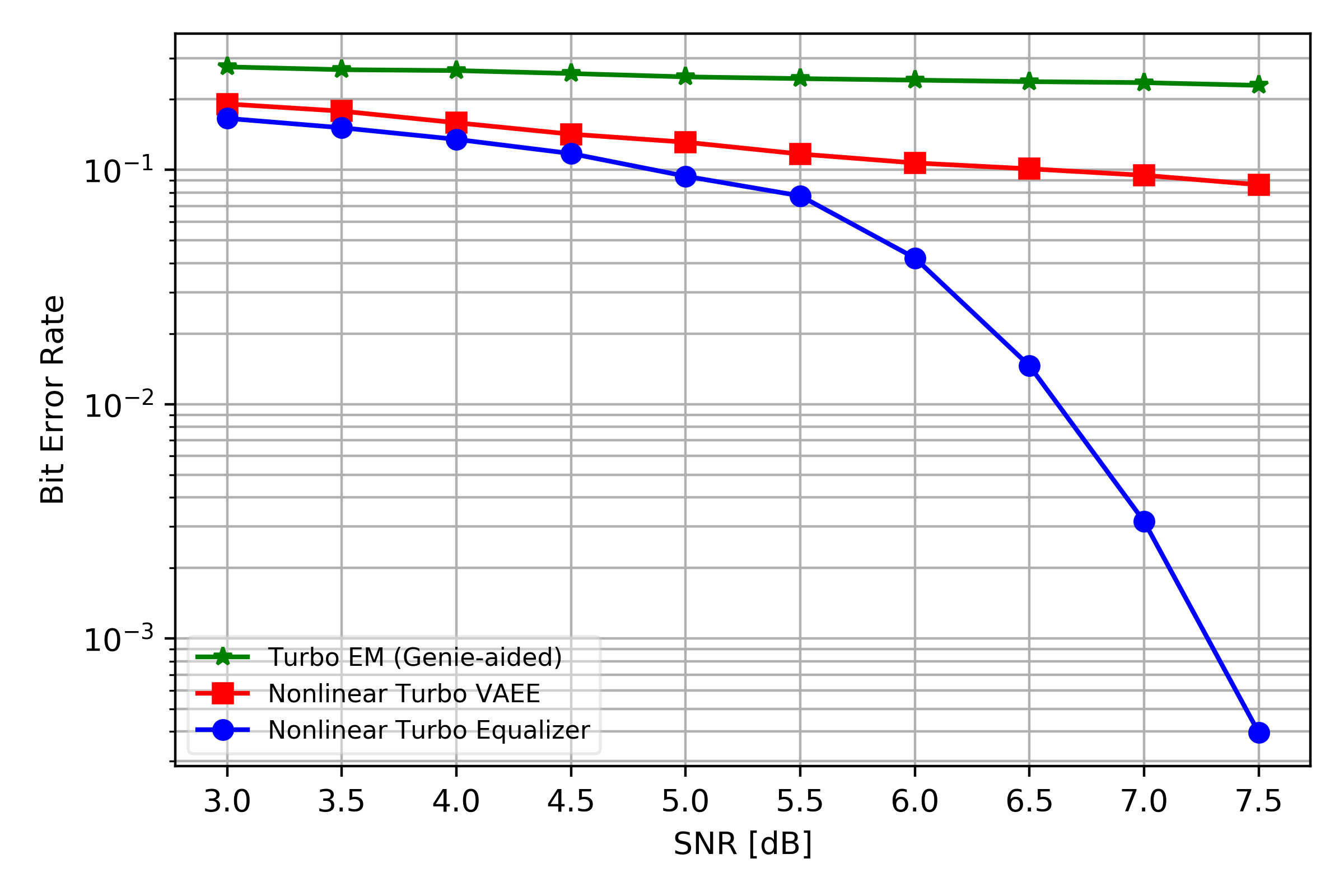}
		\caption{LDPC code with blocklength $N_{1}$}
	\end{subfigure}
	\hfill
	\begin{subfigure}[h]{0.49\linewidth}
		\includegraphics[width=1\linewidth]{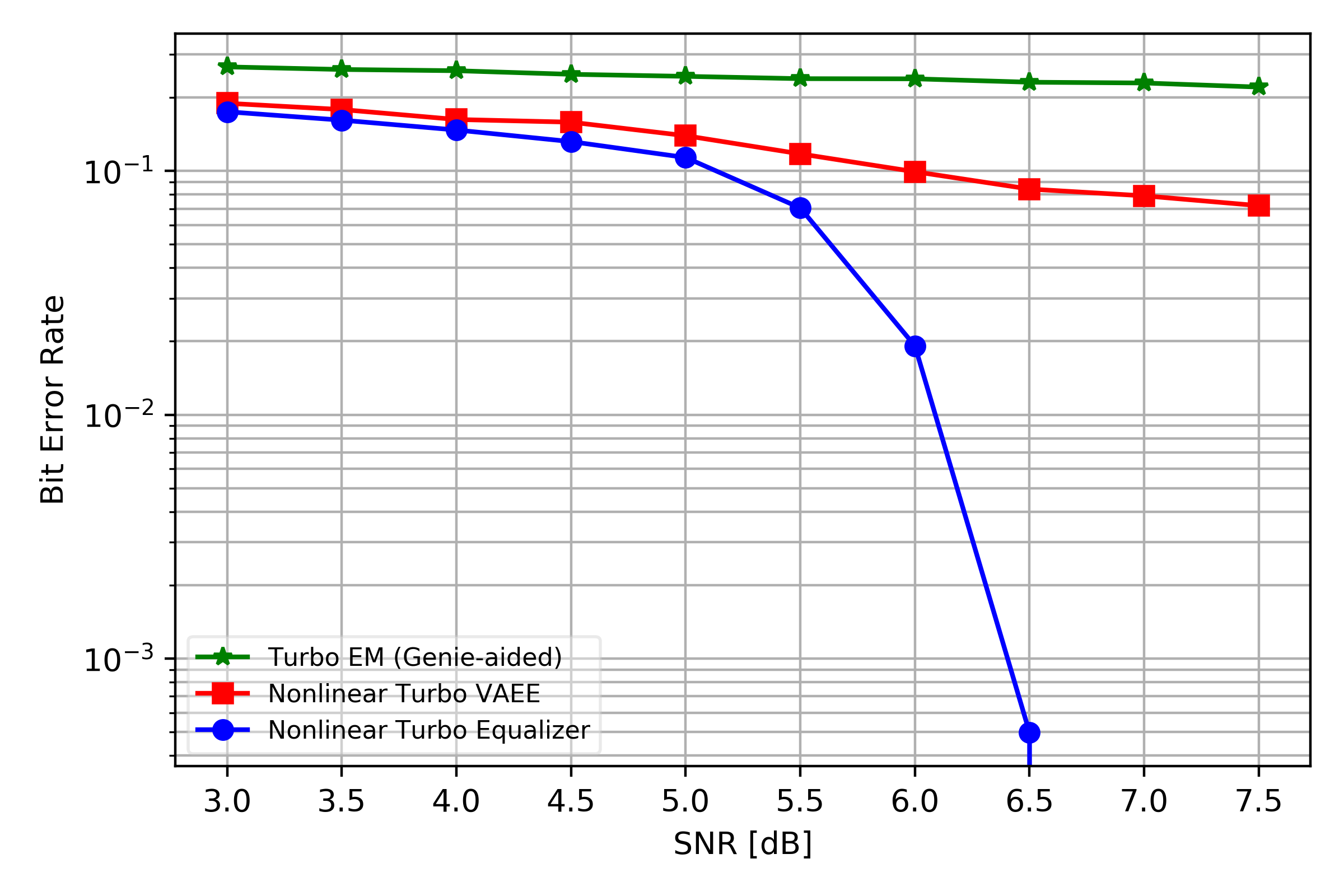}
		\caption{LDPC code with blocklength $N_{2}$}
	\end{subfigure}
	\caption{BER vs. SNR for the blind decoding algorithms and for the channel informed turbo equalizer. The channel is $\tilde{\bh}_3$. The nonlinearity is $g_{3}(\cdot)$.}
	\label{fig:snr_g3}
\end{figure}

In the following simulations, we used the channel $\tilde{\bh}_{3}$, and the same two LDPC codes with blocklengths $N_{1}=576$ and $N_{2}=2304$ that were used in the previous section.
The dropout probabilities in Fig. \ref{fig:encoderA}, describing the neural network $A$, were $0.3$.
The results are presented in Figs. \ref{fig:snr_g1}, \ref{fig:snr_g2} and \ref{fig:snr_g3}.
The SNR is now defined by
$
\text{SNR} \defined 20\log_{10} \left( ||\bg(\bx*\bh)|| / ||\bw|| \right)
$.
As can be seen, the new iterative turbo VAEE algorithm significantly outperforms the blind turbo genie-aided EM algorithm. This is not surprising since the turbo EM algorithm ignores the nonlinearity, as mentioned above.

\section{Conclusion} \label{sec:conclusion}
We introduced novel unsupervised neural network-based algorithms for blind channel equalization using the method of variational autoencoders. Both linear and nonlinear noisy ISI channels were considered. The results were then extended to joint equalization and decoding of LDPC codes using an iterative turbo VAEE algorithm. We showed significantly improved BER performance compared to the baseline algorithms. For LDPC coded data, only a genie-aided turbo EM algorithm performed well for some of the channels, and even this genie-aided implementation did not work well for the longer impulse response.
Furthermore, the computational complexity of turbo EM is exponentially increasing in the length of the estimated channel impulse response, since it uses a trellis-based equalizer, where the number of states grows exponentially with the length of the estimated impulse response. Turbo VAEE, on the other hand, uses a simple convolutional neural network.
Future research should extend our method to generalized setups such as higher constellations and channel acquisition for massive MIMO.


\bibliographystyle{IEEEtran}
\bibliography{mybib}

\end{document}